\documentclass[lettersize,journal]{IEEEtran}
\usepackage{amsmath,amssymb}
\usepackage{algorithmic}
\usepackage{algorithm}
\usepackage{array}
\usepackage{textcomp}
\usepackage{stfloats}
\usepackage{url}
\usepackage{verbatim}
\usepackage{graphicx}
\makeatletter
\let\NAT@parse\undefined
\makeatother
\usepackage[numbers,sort]{natbib}

\usepackage{bm}
\usepackage{xcolor}
\usepackage{colortbl}
\usepackage{mathtools}
\usepackage{hyperref}
\usepackage{bookmark}%
\usepackage{cleveref}
\usepackage{siunitx}
\usepackage{booktabs}
\usepackage{etoolbox}
\usepackage{tikz}
\usetikzlibrary{shapes}
\usepackage{multirow}
\usepackage{makecell}
\usepackage{cuted}
\usepackage{gensymb}

\usepackage[caption=false,font=normalsize,labelfont=sf,textfont=sf]{subfig}
\usepackage{tablefootnote}

\definecolor{figblue}{HTML}{2a78d6}
\definecolor{figorange}{HTML}{eb6834}
\definecolor{figgreen}{HTML}{1baf7a}
\definecolor{figamber}{HTML}{eda100}
\definecolor{figpink}{HTML}{e87ba4}
\definecolor{figgray}{HTML}{8a8984}
\newcommand{\tabbest}[1]{\cellcolor{figblue!30}$\bm{#1}$}
\newcommand{\tabsecond}[1]{\cellcolor{figblue!12}$#1$}

\Crefformat{figure}{#2Fig.~#1#3}
\Crefmultiformat{figure}{Figs.~#2#1#3}{ and~#2#1#3}{, #2#1#3}{ and~#2#1#3}
\Crefformat{equation}{#2Eq.~#1#3}

\usepackage{xspace}
\makeatletter
\DeclareRobustCommand\onedot{\futurelet\@let@token\@onedot}
\def\@onedot{\ifx\@let@token.\else.\null\fi\xspace}
\makeatother

\newcommand{\method}{FMP\xspace}

\newcommand{\mw}{\bm{x}}
\newcommand{\noise}{\bm{\epsilon}}
\newcommand{\cond}{\bm{c}}
\newcommand{\vel}{\bm{v}_\theta}
\newcommand{\mwref}{\hat{\bm{x}}}
\newcommand{\tproj}{\tau_s}

\newcommand{\norm}[1]{\left\lVert#1\right\rVert}

\usepackage{pifont}%

\def\thickhline{\noalign{\hrule height.8pt}}
\makeatletter
\patchcmd{\@makecaption}{\par\addvspace{0.5\baselineskip}\egroup}{\par\addvspace{0.25\baselineskip}\egroup}{}{}
\makeatother
\newcommand{\tabnote}[1]{\par\vskip\abovecaptionskip\parbox[t]{\linewidth}{\normalfont\footnotesize\noindent #1}\par}

\definecolor{black}{HTML}{000000}
\definecolor{orange}{HTML}{E69F00}
\definecolor{lightblue}{HTML}{56B4E9}
\definecolor{green}{HTML}{009E73}
\definecolor{yellow}{HTML}{F0E442}
\definecolor{blue}{HTML}{0072B2}
\definecolor{red}{HTML}{D55E00}
\definecolor{pink}{HTML}{CC79A7}

\usepackage{fancyhdr}
\begin{document}
\bstctlcite{IEEEexample:BSTcontrol}

\title{FMP: Flow-Matching Motion Prior for Humanoid Control}

\author{
Jin Cheng, Fatemeh Zargarbashi, Yijiang Huang, Taerim Yoon, Stelian Coros%
\thanks{Computational Robotics Lab at ETH Zurich, Switzerland}%
\thanks{Project website: \url{https://fmp-humanoid.github.io/}}%
}

\maketitle

\begin{abstract}

Rewards for a whole-body humanoid skill are often defined by its outcome, but such rewards do not determine the intermediate motion that leads to it.
A tennis return, for example, can be rewarded by where the ball lands, leaving the approach, the preparation of the swing, and the follow-through unspecified.
Our goal is to use human motion data to provide a motion prior that explicitly guides the policy toward movements that are both natural and appropriate to the task at each moment.
We propose Flow-Matching Motion Prior (\method), a motion prior that guides policy training with an explicit, task-conditioned reference, obtained by projecting the robot's motion onto the data manifold of a pretrained flow-matching model.
Unlike in motion tracking, no clip has to be selected or time-aligned, and unlike with adversarial and score-matching priors, the reward has an explicit reference, so it can be inspected, tuned term by term, and restricted to parts of the body.
We evaluate \method on locomotion, stylized locomotion, box carrying, and tennis, both in simulation and on a real Unitree G1 humanoid.
In tennis, a single policy trained with the task-conditioned prior learns both forehand and backhand strokes, while neither an unconditional prior nor the task reward alone achieves this.
All policies transfer to the robot without fine-tuning and use neither a reference motion nor the prior at deployment.

\end{abstract}


\section{Introduction}
\label{sec:introduction}

\IEEEPARstart{H}{umanoid} robots are designed to perform skills that involve their whole body, such as moving to a target location~\citep{radosavovic2024locomotion}, performing dance movements~\citep{cheng2024exbody,he2025asap,liao2025beyondmimic}, transporting objects~\citep{dao2024box,yang2025omniretarget,zhao2025resmimic}, or even playing dynamic sports~\citep{haarnoja2024soccer,su2025hitter,zhang2026latent}.
In reinforcement learning (RL), the outcome of such a skill, such as placing a box at a goal or landing a tennis ball at a target, is straightforward to reward, but these task-related rewards do not determine the motion that achieves it.
Natural and appropriate movements, such as reaching, lifting, or preparing a stroke and following through when returning a ball, typically require careful, task-specific reward engineering.
We aim to reduce this burden by providing a motion prior that rewards the policy for movements similar to human demonstrations and appropriate to the task context.

There are two main approaches that bring human demonstrations into policy training.
The most direct one is to track reference motions~\citep{peng2018deepmimic,liao2025beyondmimic,liao2026supermimic,luo2025sonic}.
The reference is explicit, and the tracking terms are intuitive and easy to tune, but the motion fits the task only if a suitable clip is selected and aligned in time.
Learned motion priors instead score how well the policy's motion matches the movement patterns of the data, without prescribing a trajectory, and leave the policy free to choose its motion according to the task reward.
Adversarial~\citep{peng2021amp} and score-matching~\citep{mu2025smp} methods learn this score from the data and keep the motion close to it, but they cannot provide an explicit reference that would indicate which movement fits the task at a given moment.
For example, a well-formed swing at the wrong time is natural, but it does not land the ball at the target.

\begin{figure}
    \centering
    \includegraphics[width=\linewidth]{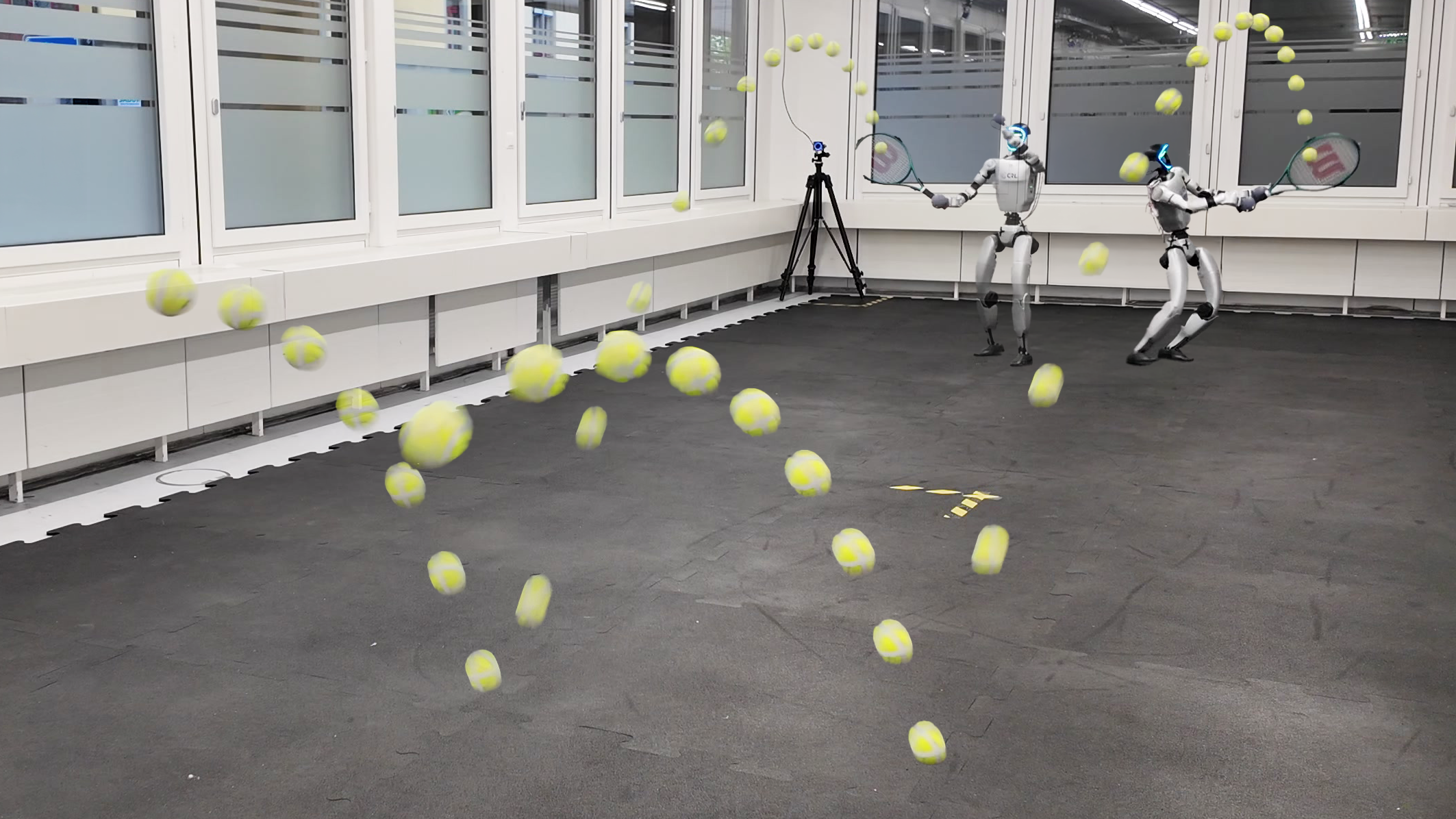}
    \caption{A policy trained with \method plays tennis autonomously on the Unitree G1, returning incoming balls with both forehand and backhand strokes, using neither a reference motion nor the prior at deployment. Overlaid poses and ball positions show the strokes and the ball trajectories.}
    \label{fig:teaser}
    \vspace{-0.2cm}
\end{figure}

To address these challenges, we introduce \emph{Flow-Matching Motion Prior} (\method), a motion prior that steers the policy toward motion that is both close to the data and appropriate to the current task context, in the form of an explicit reference that the policy can track.
It builds on a pretrained flow-matching model of kinematic motion, conditioned on the task context, such as the hitting target and the time to contact in tennis.
Inspired by SDEdit~\citep{meng2022sdedit}, \method \emph{projects} the robot's motion toward the prior manifold by partially noising the motion and integrating the conditioned flow.
\method guides policy learning in sparse-reward settings using a reference derived from the policy's own motion conditioned on the task context, rather than a selected and time-aligned clip.
With \method, a single policy learns both forehand and backhand tennis strokes from a task reward given only around contact (\Cref{fig:teaser}).

We validate these properties on tasks including locomotion, stylized locomotion, box carrying, and tennis.
We first show that the projection, the core mechanism behind \method, corrects unnatural motions produced during training toward the data manifold while preserving their overall structure (\Cref{sec:exp_projection}).
With the task condition, the projection of \method also improves task success over both unconditional priors and the task reward alone, while keeping the motion close to the data (\Cref{sec:exp_properties}).
In addition, we show that one frozen prior serves several tasks (\Cref{sec:exp_prior}) and that having the explicit reference allows the style reward to be adapted to parts of the body without retraining the prior (\Cref{sec:exp_partial}).
We further demonstrate that the projection time sets how strongly the prior guides the policy and can be adjusted to the task (\Cref{sec:exp_design}), and all policies transfer to a real Unitree G1 humanoid without fine-tuning (\Cref{sec:exp_deploy}).

Our contributions are summarized as follows:
\begin{itemize}
    \item We propose \method, a motion prior that guides policy training with an explicit reference, obtained by projecting the policy's motion onto the data manifold of a pretrained flow-matching model.

    \item We show that task conditioning of \method results in motion that is close to the data and appropriate to the current task.

    \item We further demonstrate that \method is reusable across tasks with one frozen prior and across different body parts through its explicit reference.

    \item We deploy the policies trained with \method on real hardware in a zero-shot manner.

\end{itemize}

\section{Related Work}
\label{sec:related}

Human motion data enters the training of humanoid controllers either as an explicit reference that the policy tracks, or as a learned prior over motions that shapes the reward.
We review both directions, as well as work on humanoid sports.

\subsection{Motion Tracking}
\label{sec:rw_tracking}

The most direct way to obtain human-like motion is to reward a policy for following a reference clip~\citep{peng2018deepmimic}.
Trained on large motion datasets, a single policy can track a wide range of motions~\citep{luo2023phc,liao2026supermimic}.
On humanoid hardware, tracking has enabled expressive and highly dynamic whole-body motion~\citep{cheng2024exbody,he2025asap,chen2025gmt,liao2025beyondmimic,luo2025sonic} as well as whole-body teleoperation~\citep{he2024h2o,he2024omnih2o,fu2024humanplus,ze2025twist}.
If the reference additionally contains the scene or a manipulated object, the same principle yields scene interaction and loco-manipulation skills~\citep{allshire2025videomimic,weng2025hdmi,zhao2025resmimic,yang2025omniretarget}.

A tracking policy typically needs a reference at run time.
For autonomous behavior, the reference can be produced by a kinematic generative model, typically a motion diffusion model~\citep{tevet2023mdm}, whose output is tracked in simulation, during denoising~\citep{yuan2023physdiff}, in closed loop~\citep{tevet2025closd}, or on a robot~\citep{xie2026textop}.
Alternatively, a tracking policy can be distilled into a generative controller to remove the need for a reference during inference.
This can be a latent-variable policy whose latent space serves as the action space of a task policy~\citep{luo2024pulse} or which is prompted with partial goals~\citep{tessler2024maskedmimic}, or a diffusion policy~\citep{truong2024pdp} that is steered toward a task by guidance at test time~\citep{huang2025diffusecloc,liao2025beyondmimic}.
Our proposed \method also has the form of a tracking reward.
Its reference, however, is not given to the policy.
It is computed during training from the motion of the policy itself, and the trained policy receives only the task command at inference time.

\subsection{Learned Motion Priors}
\label{sec:rw_priors}

Instead of prescribing a trajectory, a learned motion prior rewards the policy for moving in a way that is likely under a motion dataset, leaving the choice and timing of the motion to the task reward.
Adversarial Motion Priors (AMP)~\citep{peng2021amp} obtains such a style reward from a discriminator that is trained to distinguish the motion of the policy from the data, and combines it with task rewards.
Adversarial priors have been extended with latent variables to reusable skill embeddings~\citep{peng2022ase,tessler2023calm}, transferred to legged robots~\citep{escontrela2022amp,vollenweider2023multiamp}, and used for natural locomotion of humanoid robots~\citep{zhang2024adam,alvarez2025costume}.
For simulated characters, they have also been applied to interaction.
InterPhys~\citep{hassan2023interphys} conditions the discriminator and the policy on the scene and learns carrying, sitting, and lying down, and later works compose part-wise priors~\citep{bae2023pmp} or extend carrying to several characters~\citep{gao2024coohoi}.
In all these methods, the discriminator is trained together with the policy, so the prior has to be retrained for every policy and needs access to the whole motion dataset during policy training.

Recent work replaces the discriminator with a generative model that is trained on the motion data in advance.
NEAR~\citep{diwan2025near} learns noise-conditioned energy functions by denoising score matching and uses them as rewards.
Score-Matching Motion Priors (SMP)~\citep{mu2025smp} pretrains a motion diffusion model and derives a reward from it by score distillation, which makes the frozen model reusable across tasks; diffusion-based style rewards have also been used for quadrupedal locomotion~\citep{chen2026diffcast}.
\method belongs to this group; in addition, its reward tracks an explicit reference motion projected from the policy's recent trajectory, which can be inspected, tuned, and restricted to parts of the body, and its prior can be conditioned on the task context.

\subsection{Humanoid Sports}
\label{sec:rw_sports}

Sports require whole-body motion that is timed to a fast-moving object.
They were first studied with simulated characters, in football~\citep{liu2022football}, tennis~\citep{zhang2023vid2player3d}, and table tennis~\citep{wang2024tabletennis}, and have since been demonstrated on hardware, in soccer~\citep{haarnoja2024soccer,wang2025soccer,ren2025goalkeeper}, table tennis~\citep{su2025hitter,hu2025tabletennis}, badminton~\citep{liu2025badminton,chen2026badminton}, including on a legged manipulator~\citep{ma2025badminton}, and tennis~\citep{zhang2026latent,huang2026adapt}.
These works differ mainly in where the motion of the body comes from, and we group them accordingly.
Some systems use no motion data and shape the motion with task rewards and training curricula~\citep{hu2025tabletennis,ma2025badminton,liu2025badminton}.
Others combine a planner, which selects or generates the stroke for the incoming ball, with a controller that imitates human motion~\citep{zhang2023vid2player3d,wang2024tabletennis,su2025hitter,huang2026adapt}, or correct and compose fragments of human motion~\citep{zhang2026latent}.
Among them, Vid2Player3D~\citep{zhang2023vid2player3d} is the closest to our tennis task in its data, since its strokes are likewise reconstructed from video.
HumanX~\citep{wang2026humanx} imitates interaction skills, such as basketball and football, directly from human videos.
A third group regularizes the policy with adversarial priors~\citep{wang2025soccer,ren2025goalkeeper,chen2026badminton}, which can be conditioned on the ball position~\citep{ren2025goalkeeper}.

Our tennis task belongs to the last group, where a single policy is trained with a task reward concentrated around the ball contact and a motion prior conditioned on the hitting target.
The prior guides the stroke to coordinate its timing with the incoming ball, and at deployment the policy executes the entire stroke without a separate module that selects, generates, or tracks a reference motion.

\section{Background: Flow Matching}
\label{sec:background}

Flow matching~\citep{lipman2023flow,liu2023flow} learns to generate samples from a data distribution by gradually transforming Gaussian noise into data.
We denote a data sample by $\mw^1$ and a noise sample of the same shape by $\noise \sim \mathcal{N}(\bm{0}, \bm{I})$; superscripts refer to the generative process.
Noise and data are connected by a probability path that is indexed by a continuous flow time $\tau \in [0, 1]$.
We use the linear path
\begin{equation}
    \mw^{\tau} = (1 - \tau)\, \noise + \tau\, \mw^1 ,
    \label{eq:path}
\end{equation}
so that $\tau = 0$ corresponds to pure noise and $\tau = 1$ to data.
Along this path, a sample moves with the constant velocity $\mw^1 - \noise$, and a velocity field $\vel$ is trained to predict it,
\begin{equation}
    \mathcal{L}(\theta) = \mathbb{E}_{\mw^1,\; \noise,\; \tau \sim \mathcal{U}[0, 1]}\, \norm{\vel(\mw^{\tau}, \tau) - (\mw^1 - \noise)}^2 .
    \label{eq:fm_objective}
\end{equation}
Many pairs of data and noise samples pass through the same point $\mw^{\tau}$, and the minimizer of \Cref{eq:fm_objective} is their mean velocity.
The ordinary differential equation $\mathrm{d}\mw / \mathrm{d}\tau = \vel(\mw, \tau)$ then transports the noise distribution at $\tau = 0$ to the data distribution at $\tau = 1$~\citep{lipman2023flow}.
Samples are generated by integrating it from a noise sample with Euler steps of size $\Delta\tau$,
\begin{equation}
    \mw^{\tau + \Delta\tau} = \mw^{\tau} + \Delta\tau \; \vel(\mw^{\tau}, \tau) .
    \label{eq:euler}
\end{equation}
The linear path often leads to straighter trajectories than those of diffusion models, so comparatively few integration steps suffice~\citep{liu2023flow}.
Since the velocity predicts $\mw^1 - \noise$, solving \Cref{eq:path} for the data sample gives a one-step estimate of the clean sample, $\mw^{\tau} + (1 - \tau)\, \vel(\mw^{\tau}, \tau)$, which is one Euler step from $\tau$ to $1$.

\section{Method}
\label{sec:method}

\begin{figure}[t]
    \centering
    \includegraphics[width=\linewidth]{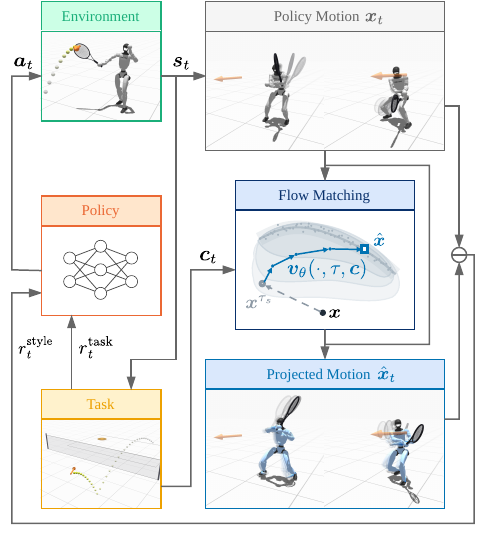}
    \caption{Overview of \method. The last $H$ frames of the robot form the policy window $\mw_t$. The frozen flow-matching prior blends it with noise up to the projection time $\tproj$ and integrates the flow $\vel$ back to $\tau = 1$ under the task condition $\cond_t$, which yields the reference window $\mwref_t$. The style reward $r_t^{\text{style}}$ tracks the residual between the two windows and is combined with the task reward $r_t^{\text{task}}$. Only the prior sees the motion data.}
    \label{fig:overview}
\end{figure}

\method leaves \emph{what to achieve} to the task reward and takes \emph{how to move}, given the task context, from a prior learned on motion data.
First, we train a flow-matching generative model on short windows of reference motion.
This model is the only component that ever sees the motion data.
In a second stage, the model is frozen and used as a motion prior for RL.
At every control step, our method projects the recent motion of the robot onto a nearby motion on the data manifold, and the policy is rewarded for staying close to this reference (\Cref{fig:overview}).
When the prior is conditioned on the task context, the same reward also favors motions that fit this context.
\Cref{sec:policy} describes how the resulting style reward is combined with task rewards and how the prior is used to initialize episodes.
\Cref{sec:smp} relates our reward to other motion priors, in particular to SMP~\citep{mu2025smp}, the approach closest to ours.

\subsection{Motion Representation}
\label{sec:representation}

We represent motion as a window of $H$ consecutive frames at the control rate of \SI{50}{\hertz}, $\mw = [\bm{f}_1, \dots, \bm{f}_H] \in \mathbb{R}^{H \times D}$.
Each frame $\bm{f}_h \in \mathbb{R}^{D}$ contains the root position, the root orientation in the continuous 6D representation~\citep{zhou2019continuity}, the joint positions, the positions of key bodies, namely the two feet, the torso, and the two wrists, and the root linear and angular velocities (\Cref{tab:features} in Appendix~\ref{app:priors}).

To make the representation invariant to the global position and heading of the robot, all quantities of a window are expressed in a canonical frame defined by its \emph{last} frame. 
The origin is the horizontal root position of the last frame and the $x$-axis is aligned with its heading.
The root height is kept in world coordinates, and key-body positions are taken relative to the root of their own frame.
We use $H = 10$ frames (\SI{0.2}{\second}).
Each feature dimension is normalized to $[-1,1]$.
All windows are normalized unless stated otherwise.

\subsection{Flow-Matching Motion Prior}
\label{sec:prior}

Given a dataset $\mathcal{D}$ of motion windows, optionally paired with a condition $\cond$, we model the distribution $p(\mw \mid \cond)$ with flow matching.
A data sample $\mw^1$ is a normalized motion window, and the velocity field $\vel$ additionally receives the condition.
We train it on the linear path of \Cref{eq:path}, 
\begin{equation}
    \mathcal{L}(\theta) = \mathbb{E}_{(\mw^1, \cond) \sim \mathcal{D},\; \noise,\; \tau \sim \mathcal{U}[0, 1]}
    \norm{\vel(\mw^{\tau}, \tau, \cond) - (\mw^1 - \noise)}_1 .
    \label{eq:fm_loss}
\end{equation}
New motion windows are generated from Gaussian noise by integrating the flow from $\tau = 0$ to $\tau = 1$ with Euler steps.

The velocity field is a small transformer that operates on the $H$ frames of a window.
Each frame is linearly embedded into a token of width 128 and combined with a sinusoidal positional encoding, followed by two transformer blocks with four-head self-attention.
The flow time is embedded and injected into every block through adaptive layer normalization~\citep{peebles2023dit,chen2024pixart}; the condition, if present, is linearly projected and added to the time embedding.
The model has approximately 0.7M parameters.
The hyperparameters are chosen to allow fast inference, because the prior is evaluated several times per control step for thousands of parallel environments during policy training.
Training windows are sampled uniformly over time across the dataset. All training hyperparameters are listed in Appendix~\ref{app:priors}.

\subsection{Style Reward from Prior Projection}
\label{sec:style_reward}

The prior defines which motions are natural.
The remaining question is how to turn it into a reward for a policy whose motion is, especially early in training, far from the data.
We take inspiration from SDEdit~\citep{meng2022sdedit}, which edits an image by partially noising it and then denoising it with a pretrained generative model, so that the result keeps the coarse structure of the input while becoming realistic.
We use the frozen prior in a similar way to \emph{project} the recent motion of the policy onto the data manifold, and reward the policy for staying close to its projection.

\subsubsection{Policy window}
During rollouts, each environment keeps a buffer with the last $H$ frames of the simulated robot.
At control step $t$, the buffer is converted into a motion window in the same way as the training data.
It is canonicalized with respect to the current frame and normalized.
We denote the result by $\mw_t^1$, since it takes the place of the data sample in \Cref{eq:path}; subscripts denote control steps.

\subsubsection{Projection}
We first move the policy window onto the probability path of the prior by blending it with noise,
\begin{equation}
    \mw_t^{\tproj} = (1 - \tproj)\, \noise + \tproj\, \mw_t^1 ,
    \label{eq:noising}
\end{equation}
where the \emph{projection time} $\tproj \in (0, 1)$ sets the amount of injected noise.
We then integrate the flow over the remaining interval $[\tproj, 1]$ with $N$ Euler steps of size $\Delta\tau = (1 - \tproj) / N$,
\begin{equation}
    \mw_t^{\tau_{n+1}} = \mw_t^{\tau_n} + \Delta\tau \; \vel\!\left(\mw_t^{\tau_n}, \tau_n, \cond_t\right), \qquad \tau_n = \tproj + n\, \Delta\tau ,
    \label{eq:projection}
\end{equation}
for $n = 0, \dots, N - 1$, and call the end point $\mwref_t \coloneqq \mw_t^{\tau_N}$ the \emph{reference window} ($\tau_N = 1$).

The injected noise covers small deviations of the policy motion, while its overall structure, such as the posture, the direction of travel, and the phase of the movement, remains visible, and the flow regenerates what the noise has covered from the data distribution.
The reference window therefore resembles what the policy is currently doing but lies closer to the data manifold, and the remaining difference indicates how the motion should change.

The projection time controls this behavior.
For $\tproj \to 1$, the reference coincides with the policy window and the reward carries no information; for $\tproj \to 0$, it becomes a sample from the prior that is unrelated to what the robot is doing.
We use intermediate values $\tproj \in [0.4, 0.8]$, depending on the task, and $N = 4$ steps.

The noise $\noise$ is drawn once per environment at the beginning of an episode and kept fixed until the next reset.
Within an episode, the projection is then a deterministic function of the motion window and the reward does not fluctuate from step to step due to resampling, while different episodes and environments still cover different noise samples.

\subsubsection{Style reward}
Let $\bm{\delta}_t = \mwref_t - \mw_t^1 \in \mathbb{R}^{H \times D}$ be the residual between the reference and the policy window.
We partition the feature dimensions into the six groups $\mathcal{G}_g$ of \Cref{tab:features} and compute one error per group,
\begin{equation}
    e_{g,t} = \frac{1}{H D} \sum_{h=1}^{H} \sum_{d \in \mathcal{G}_g} \delta_{t,h,d}^{2} .
    \label{eq:group_error}
\end{equation}
The style reward is the product of one kernel per group,
\begin{equation}
    r_t^{\text{style}} = \prod_{g} \exp\!\left( - \kappa\, w_g\, \frac{e_{g,t}}{\bar{e}} \right) \in (0, 1],
    \label{eq:style_reward}
\end{equation}
with sharpness $\kappa$, group weights $w_g$, and scale $\bar{e}$.
The magnitude of the residual depends on the prior and on $\tproj$.
We therefore divide by the running mean $\bar{e}$ of the total error $\sum_g e_{g,t}$ over all environments and steps.
The running mean is accumulated during an initial phase of training and then frozen, so that the reward is stationary afterwards.
With this scale, the same $\kappa$ transfers across priors and tasks.

This reward has the form of the tracking rewards used for motion imitation~\citep{peng2018deepmimic}, with two differences.
First, the reference is not a pre-recorded clip but is produced online from the motion of the policy itself, so no clip has to be selected or time-aligned and the policy remains free to choose any motion that fits its task.
Second, the reward compares entire windows, including the velocities and the evolution of the pose over the last \SI{0.2}{\second}, instead of a single frame.
Because the feature normalization is affine, every squared residual is a physical tracking error measured relative to the range that the feature covers in the data, which makes errors with different units comparable without hand-tuned scales.
The weights $w_g$ allow individual groups to be emphasized or disabled.
We set $w_g = 1$ in all tasks, for which the reward reduces to a single kernel on the mean squared residual of the window.

\Cref{alg:style_reward} summarizes the computation.
It only requires $N$ forward passes of the prior per control step, batched over all environments, and no gradients through the prior.

\begin{algorithm}[t]
    \caption{\method style reward at control step $t$}
    \label{alg:style_reward}
    \begin{algorithmic}[1]
        \REQUIRE frozen prior $\vel$, motion buffer, episode noise $\noise$, condition $\cond_t$ (optional), projection time $\tproj$, steps $N$
        \STATE $\mw^1 \leftarrow$ canonicalize and normalize the motion buffer
        \STATE $\mw \leftarrow (1 - \tproj)\, \noise + \tproj\, \mw^1$
        \STATE $\Delta\tau \leftarrow (1 - \tproj) / N$
        \FOR{$n = 0, \dots, N - 1$}
            \STATE $\mw \leftarrow \mw + \Delta\tau \; \vel(\mw,\; \tproj + n\, \Delta\tau,\; \cond_t)$
        \ENDFOR
        \STATE $\bm{\delta} \leftarrow \mw - \mw^1$
        \STATE $e_g \leftarrow \frac{1}{HD} \sum_{h} \sum_{d \in \mathcal{G}_g} \delta_{h,d}^2$ for every group $g$
        \RETURN $r^{\text{style}} = \prod_g \exp(-\kappa\, w_g\, e_g / \bar{e})$
    \end{algorithmic}
\end{algorithm}

\subsection{Task-Conditioned Priors}
\label{sec:conditional}

An unconditional prior captures how the robot should move, but not when a particular movement is appropriate.
In many tasks, the right motion depends on the context. For example, a tennis swing has to be timed to the incoming ball, and the way a box is approached depends on where it is.
For such tasks we train the prior conditionally, $\vel(\mw, \tau, \cond)$, where $\cond$ describes the context of a window.
Consistent with the motion representation, $\cond$ is defined at the last frame of the window and expressed in its heading frame, and continuous conditions are standardized.
The prior is always trained with the condition.

During policy training, the condition $\cond_t$ is computed at every control step from the simulator state and the task command and is passed to the prior during the projection.
The reference window is then a projection toward the motions that are natural \emph{in this context}.
The style reward thus also penalizes motion that would be natural in isolation but does not fit the task, for example a well-formed swing at the wrong time, and it does so without any change to the reward itself.
Any quantity that can be computed both from the motion data and from the simulator can serve as a condition; the conditions of our tasks, a style label, a hitting target, and the positions of a box and its goal, are defined in \Cref{sec:tasks}.

\subsection{Policy Learning}
\label{sec:policy}

The policy $\pi(\bm{a}_t \mid \bm{o}_t)$ maps an observation $\bm{o}_t$ of proprioception and the task command to an action $\bm{a}_t$; the reference motion is never part of its observation.
The policy is trained to maximize the expected discounted return $\mathbb{E}_{\pi} [ \sum_{t \geq 0} \gamma^{t} r_t ]$ of a reward $r_t$ that combines the style reward with a task reward $r_t^{\text{task}}$, which specifies what the robot should achieve.

\subsubsection{Combining task and style rewards}
The style reward is a bounded per-step quantity and can be combined with any task reward.
We use two schemes (\Cref{tab:tasks}).

\emph{Multiplicative gating.}
With a task reward $r_t^{\text{task}} \in (0, 1]$, the first scheme is
\begin{equation}
    r_t = r_t^{\text{task}} \cdot \left( \beta + (1 - \beta)\, r_t^{\text{style}} \right) \cdot r_t^{\text{reg}} ,
    \label{eq:gated_reward}
\end{equation}
where $r_t^{\text{reg}} = \exp(-\norm{\bm{a}_t - \bm{a}_{t-1}}^2 / \sigma_a^2)$ regularizes the action rate.
Due to the product, task progress only pays off when it is achieved with natural motion.
The floor $\beta \in [0, 1)$ limits how much of the task reward the prior can withhold, which preserves a learning signal while the motion of the policy is still far from the data.

\emph{Multi-critic.}
The second scheme treats $r^{\text{task}}$, $r^{\text{reg}}$, and $r^{\text{style}}$ as separate reward channels with one value function each~\citep{zargarbashi2024robotkeyframing,cheng2024diverse}, a form of multi-critic learning~\citep{mysore2022multicritic}.
The advantages $\hat{A}_t^{i}$ are estimated per channel with generalized advantage estimation (GAE)~\citep{schulman2016gae}, standardized, and then combined as
\begin{equation}
    \hat{A}_t = \sum_{i \in \{\text{task},\, \text{reg},\, \text{style}\}} \lambda_i\, \hat{A}_t^{i} ,
    \label{eq:multi_critic}
\end{equation}
which decouples the relative importance $\lambda_i$ of the channels from the scale and density of their rewards.

\subsubsection{Generative state initialization}
\label{sec:gsi}
Following~\citet{mu2025smp}, we initialize episodes from motion windows generated by the prior instead of from the motion dataset, so that the frozen prior is the only interface between the data and the policy.
The generated window also fills the motion buffer, so that the style reward is defined from the first step.
For a conditional prior, the window is generated for the condition of the sampled task command, and task objects such as the box are placed consistently with it.

\subsection{Relation to Other Motion Priors}
\label{sec:smp}

\begin{figure}[t]
    \centering
    \includegraphics[width=\columnwidth]{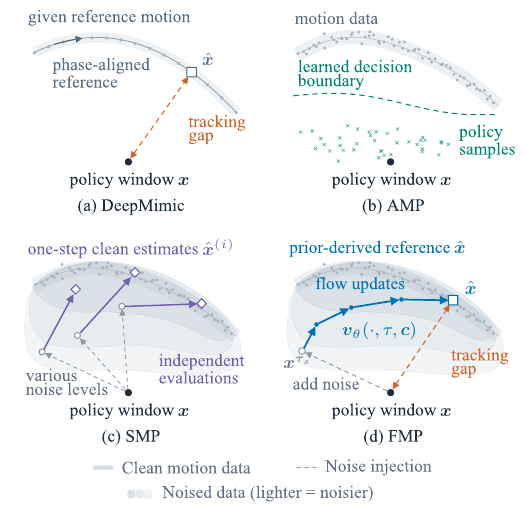}
    \caption{Sources of the style reward. The gray band is the motion data, the lighter bands its noised versions. (a) Motion tracking rewards the gap to a phase-aligned frame of a reference clip. (b) AMP rewards the policy through a discriminator trained online against policy samples. (c) SMP noises the policy window at several levels and penalizes the noise-prediction errors of the frozen model, which equal errors to one-step clean estimates. (d) \method noises the policy window once, integrates the flow of the frozen prior, and rewards the gap to the resulting reference.}
    \label{fig:priors}
\end{figure}

\begin{figure*}[t]
    \centering
    \includegraphics[width=\textwidth]{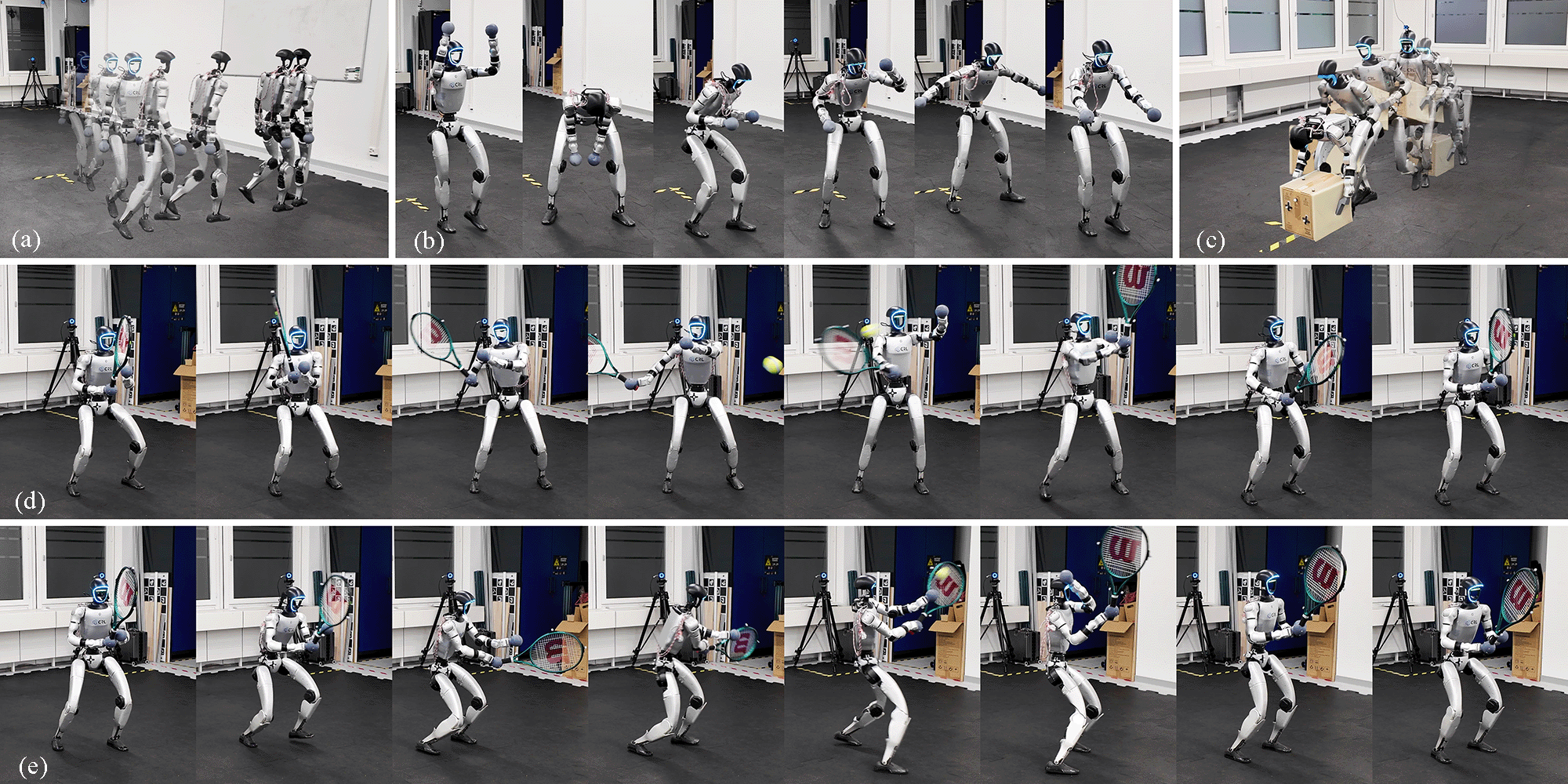}
    \caption{\method policies on the Unitree G1. (a) Goal reaching with the neutral-walking prior, poses sampled along the walk to the goal. (b) Stylized locomotion with the style-conditioned walking prior, one policy under six style commands: \emph{ArmsAboveHead}, \emph{HandsBetweenLegs}, \emph{Dinosaur}, \emph{ShieldedLeft}, \emph{Star}, and \emph{Zombie}. (c) Box carrying with the box-conditioned prior: approach, lift, carry, and set-down. (d) A forehand and (e) a backhand with the hit-conditioned tennis prior, from the ready posture through preparation, contact, and follow-through.}
    \label{fig:hardware}
    \vspace{6pt}
    \makeatletter\def\@captype{table}\makeatother
    \caption{Tasks.}
    \label{tab:tasks}
    \centering
    \footnotesize
    \begin{tabular}{>{\raggedright\arraybackslash}p{2.6cm} >{\raggedright\arraybackslash}p{5.2cm} >{\raggedright\arraybackslash}p{2.6cm} >{\raggedright\arraybackslash}p{3.0cm} >{\raggedright\arraybackslash}p{2.2cm}}
        \thickhline
        Task & Motion data & Prior condition $\cond$ & Task command & Reward \\
        \hline
        Locomotion & locomotion clips from LAFAN1~\citep{harvey2020lafan} or 100STYLE~\citep{mason2022style}, per experiment & -- & goal position or steering command & gated \\
        Stylized locomotion & 100STYLE~\citep{mason2022style}, 100 styles, \SI{6.7}{\hour} & style label & goal position, style & gated \\
        Box carrying & OmniRetarget~\citep{yang2025omniretarget}, 56 clips, \SI{5}{\minute} & box and goal position & goal location & multi-critic \\
        Tennis & 500 strokes from video, \SI{24}{\minute} & hitting target & landing target & multi-critic \\
        \thickhline
    \end{tabular}
    \tabnote{The duration of the motion data includes mirrored clips. Gated and multi-critic refer to \Cref{eq:gated_reward} and \Cref{eq:multi_critic}.}
\end{figure*}

\Cref{fig:priors} places \method among the main sources of a style reward: direct tracking of reference clips (DeepMimic)~\citep{peng2018deepmimic}, Adversarial Motion Priors~\citep{peng2021amp}, and Score-Matching Motion Priors~\citep{mu2025smp}, the approach closest to ours.
\method shares its reward form with motion tracking and its reward source, a pretrained and frozen generative model, with SMP.
SMP rewards a small noise-prediction error of its frozen diffusion model at three fixed noise levels.
It does not denoise the policy window, but its reward can equivalently be read as a tracking reward on one-step estimates of the clean window, of which our projection with a single step, $N = 1$, is the flow-matching counterpart (Appendix~\ref{app:smp}).
The two differ in how the estimate is formed. 
For an optimal model, a one-step estimate is the mean of all data windows compatible with the noised window, whereas \method integrates the flow with $N$ steps to a single motion, which yields one explicit reference per control step whose group errors can be inspected and weighted individually.
Finally, SMP conditions its prior on a motion style through classifier-free guidance~\citep{ho2022cfg}, whereas we use the conditional velocity field directly in the projection, with a task context that changes at every control step.
These differences concern the form of the reward rather than its quality.
In our comparison (\Cref{sec:exp_properties}), the two rewards keep the motion at a similar distance to the data.

\section{Experiments}
\label{sec:experiments}

We apply \method to four tasks on the Unitree G1 humanoid with 29 actuated joints (\Cref{tab:tasks} and \Cref{fig:hardware}): locomotion, stylized locomotion, box carrying, and tennis.
The motion representation, the architecture of the prior, and the style reward are identical across all tasks.
We first examine whether \method corrects noisy motions produced during exploration by projecting them onto the data manifold (\Cref{sec:exp_projection}). We then train policies to evaluate task success and motion quality (\Cref{sec:exp_properties}). We also demonstrate that \method is reusable across tasks and across parts of the body (\Cref{sec:exp_reuse}).
Finally, we study the projection time and the number of integration steps (\Cref{sec:exp_design}) and deploy the policies on hardware (\Cref{sec:exp_deploy}).

\begin{figure*}[t]
    \centering
    \includegraphics[width=\textwidth]{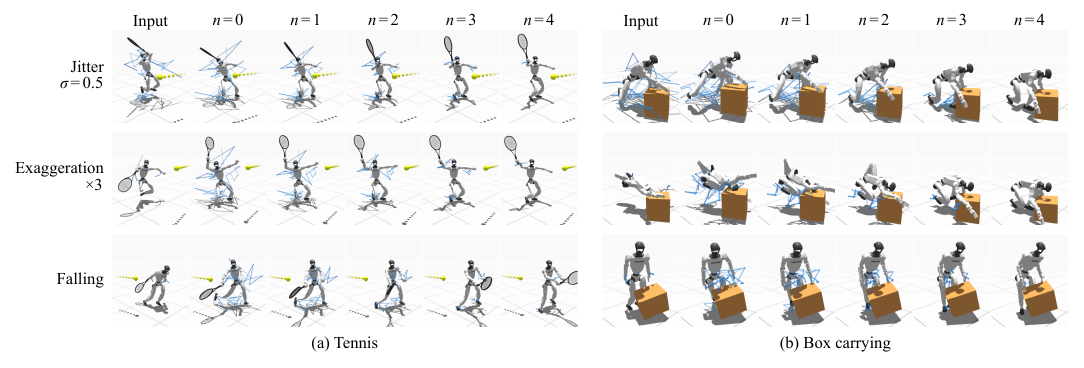}
    \caption{Motion correction through the frozen tennis and box-carrying priors. Rows: jittery motion with $\sigma=0.5$, motion exaggerated by a factor of $3$, and falling motion. Columns: the input, the noisy blend ($n=0$), and four integration steps. Each cell shows the fifth frame of its ten-frame window, with key-body trails over the whole window darkening with time. The conditioning context is drawn in each cell: in tennis, the yellow ball at the commanded contact point with a trail along the commanded outgoing direction; in box carrying, the orange box.}
    \label{fig:projection_progress}
\end{figure*}

\begin{figure}[t]
    \centering
    \includegraphics[width=\linewidth]{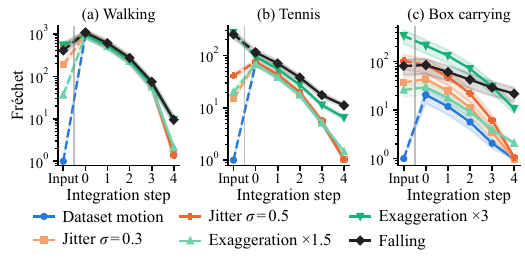}
    \caption{Fréchet distance to the motion data during projection, relative to the dataset-to-dataset distance, for (a) walking, (b) tennis, and (c) box carrying. Curves and bands show mean $\pm$ std over three seeds; the y-axes are logarithmic. Dashed segments denote the initial noising, solid segments the four integration steps.}
    \label{fig:projection_quality}
\end{figure}

\subsection{Tasks}
\label{sec:tasks}

The motion data of the tasks are described in Appendix~\ref{app:data}, and their commands, task rewards, and hyperparameters in Appendix~\ref{app:tasks}.

\subsubsection{Locomotion}
This task is the simplest setting without task conditioning, in which \method can be compared directly with other motion priors.
The policy walks to a goal or follows a steering command, and the task reward scores only the base motion or the distance to the goal, leaving the whole-body motion to the unconditional prior.

\subsubsection{Stylized locomotion}
This task tests conditioning on a discrete label, with which one prior covers many styles.
The prior is trained on the forward walks of all 100 styles of 100STYLE~\citep{mason2022style} with a one-hot style label, and one policy reaches a goal in 14 of these styles, with the style command changing during the episode.

\subsubsection{Box carrying}
This task tests conditioning on a continuous task context in loco-manipulation, with a relatively dense task reward.
The policy picks up a box of randomized size and mass, carries it to a goal \SIrange{0.8}{2}{\meter} away, and sets it down.
The prior is conditioned on the positions of the box and of the goal relative to the root, $\cond = [\bm{p}^{\text{box}}, \bm{p}^{\text{goal}}_{xy}] \in \mathbb{R}^{5}$.
The task reward depends on the box alone, leaving how the robot approaches, grasps, and carries it to the style reward.

\subsubsection{Tennis}
This task is the most demanding setting, where the task reward is sparse and the motion has to be timed to an external event.
The policy returns incoming balls over the net to a commanded target with a racket in its right hand.
The prior is trained on 500 strokes reconstructed from video and conditioned on the hitting target $\cond = [\bm{p}^{\text{hit}}, \bm{d}^{\text{hit}}, t^{\text{hit}}] \in \mathbb{R}^{7}$, where $\bm{p}^{\text{hit}}$ is the contact point relative to the root, $\bm{d}^{\text{hit}}$ the outgoing ball direction, and $t^{\text{hit}}$ the time until contact.
The task reward is given only around the contact, so the preparation and follow-through are guided by the style reward.

\subsection{Training Setup}
\label{sec:setup}

All policies are trained with Proximal Policy Optimization (PPO)~\citep{schulman2017ppo} in MuJoCo Warp through mjlab~\citep{zakka2026mjlab} with 4096 parallel environments.
The actor observes a short history of proprioceptive measurements (base angular velocity, projected gravity, joint positions and velocities, and the previous action) together with the task command, and the critic additionally receives privileged quantities such as the base linear velocity.

Actions are joint position targets tracked by PD controllers at \SI{50}{\hertz}, the frame rate of the prior.
Policy and value functions are multilayer perceptrons.
We randomize the dynamics of the robot, delay its actuation, perturb it with pushes, and add noise to its observations.
We refer readers to Appendix~\ref{app:setup} for the randomization ranges and all training hyperparameters.

\subsection{Motion Correction by Projection}
\label{sec:exp_projection}

\method corrects the noisy motions that the policy produces during exploration by projecting them toward the data manifold with the pretrained flow-matching model.
To evaluate this correction, we perturb motion windows and project them through the learned priors.
Since a corrected motion has no single ground truth, we measure how close the corrected motions are to the data distribution using the Fréchet distance (Appendix~\ref{app:frechet}).\footnote{The quantity underlying the Fréchet Inception Distance (FID)~\citep{heusel2017fid}, computed directly on the motion features rather than on network embeddings.}

To mimic the motions that a policy produces through exploration during training, we use three perturbations (rows of \Cref{fig:projection_progress}):
\begin{itemize}
    \item \emph{Jittery motion}: an explorative and under-regularized policy often yields jittery motion; we add Gaussian noise with $\sigma \in \{0.3, 0.5\}$ to every feature.
    \item \emph{Exaggerated motion}: a policy may overshoot during exploration; we scale joint angles about the default pose by $1.5$ or $3$.
    \item \emph{Falling motion}: at the start of training, the policy outputs actions that make the robot fall. We simulate this by rolling out random actions from natural initial states.
\end{itemize}
For each perturbation, we examine how much the projection corrects it.

As shown in \Cref{fig:projection_progress}, all perturbed motions become increasingly aligned with the data distribution as the integration proceeds, and their key-body trails contract into smooth paths.
In addition, \Cref{fig:projection_quality} plots the Fréchet distance of the projected motions for each perturbation over the integration steps.
Jitter, at both $\sigma = 0.3$ and $0.5$, resembles the injected noise and is almost entirely removed by the projection.
Exaggerated and falling motions remain at a higher Fréchet distance after the projection, since they have no counterpart in the data.
Rather than forcing them onto the data, the projection produces a natural motion close to the input, as shown qualitatively in \Cref{fig:projection_progress}.
This means that the reference guides the policy toward natural motion while staying close to what it is currently doing.

\begin{table*}[t]
    \caption{Comparison with other motion priors in tennis.}
    \label{tab:cond}
    \centering
    \footnotesize
    \setlength{\tabcolsep}{4pt}
    \begin{tabular}{llccccccc}
        \thickhline
        Variant & Condition & Strike $\uparrow$ & Clean hit $\uparrow$ & Cmd.\ stroke $\uparrow$ & Contact err.\ [m] $\downarrow$ & Net cleared $\uparrow$ & Landing err.\ [m] $\downarrow$ & Fréchet $\downarrow$ \\
        \hline
        Task only & -- & $0.08 \pm 0.08$ & $0.03 \pm 0.03$ & $0.02 \pm 0.03$ & $0.90 \pm 0.11$ & $0.34 \pm 0.20$ & $6.2 \pm 1.0$ & $311 \pm 160$ \\
        \hline
        AMP & uncond. & $0.03 \pm 0.00$ & $0.02 \pm 0.01$ & $0.01 \pm 0.00$ & $1.01 \pm 0.14$ & $0.06 \pm 0.04$ & $6.8 \pm 0.1$ & \tabsecond{9.0 \pm 6.3} \\
        AMP & cond. & $0.59 \pm 0.10$ & $0.41 \pm 0.08$ & $0.34 \pm 0.11$ & $0.39 \pm 0.09$ & $0.92 \pm 0.02$ & $0.82 \pm 0.10$ & \tabbest{5.6 \pm 1.7} \\
        \hline
        SMP & uncond. & $0.02 \pm 0.00$ & $0.01 \pm 0.00$ & $0.01 \pm 0.00$ & $0.72 \pm 0.02$ & $0.00 \pm 0.00$ & $8.3 \pm 0.1$ & $37.7 \pm 0.4$ \\
        SMP & cond. & \tabsecond{0.80 \pm 0.05} & \tabsecond{0.56 \pm 0.02} & \tabsecond{0.38 \pm 0.13} & \tabsecond{0.24 \pm 0.01} & \tabbest{0.98 \pm 0.00} & \tabbest{0.43 \pm 0.08} & $17.9 \pm 3.2$ \\
        \hline
        \method & uncond. & $0.07 \pm 0.01$ & $0.06 \pm 0.01$ & $0.06 \pm 0.01$ & $0.53 \pm 0.02$ & $0.01 \pm 0.01$ & $6.9 \pm 0.1$ & $32.3 \pm 2.1$ \\
        \method (ours) & cond. & \tabbest{0.81 \pm 0.08} & \tabbest{0.59 \pm 0.04} & \tabbest{0.48 \pm 0.11} & \tabbest{0.21 \pm 0.02} & \tabbest{0.98 \pm 0.02} & \tabsecond{0.44 \pm 0.05} & $14.5 \pm 0.2$ \\
        \thickhline
    \end{tabular}
    \tabnote{Mean $\pm$ std over three seeds, after the first training stage. Rates are fractions of balls, and Fréchet is relative to the dataset-to-dataset distance. Shading marks the two best means per column: dark blue and bold for the best, light blue for the second best.}
\end{table*}

\begin{table}[t]
    \caption{Effect of the task condition in box carrying.}
    \label{tab:cond_box}
    \centering
    \footnotesize
    \setlength{\tabcolsep}{4pt}
    \begin{tabular}{lccc}
        \thickhline
        Variant & Place $\uparrow$ & Dist.\ [m] $\downarrow$ & Fréchet $\downarrow$ \\
        \hline
        Task only & $0.43 \pm 0.39$ & \tabbest{0.34 \pm 0.02} & $352 \pm 67$ \\
        Uncond. & \tabsecond{0.55 \pm 0.13} & $0.71 \pm 0.14$ & \tabsecond{100 \pm 32} \\
        Cond. & \tabbest{0.83 \pm 0.01} & \tabsecond{0.38 \pm 0.03} & \tabbest{68.5 \pm 13.5} \\
        \thickhline
    \end{tabular}
    \tabnote{Mean $\pm$ std over three seeds. Place: fraction of episodes in which the box is set down at the goal; Dist.: final 3-D distance of the box to the goal; Fréchet is relative to the dataset-to-dataset distance. Dark blue and bold: best mean per column; light blue: second best.}
\end{table}

\subsection{Task Success and Motion Quality}
\label{sec:exp_properties}
We evaluate task completion and the resulting motion quality of \method on tennis and box carrying, both of which require an understanding of the task context because of the structure of the task reward.
We first compare \method with baseline methods, consisting of learned motion priors (i.e., AMP and SMP), and with task-only policies in tennis. 
Each motion prior is trained with and without the task condition.
The learned motion priors share the same expert tennis dataset and the same context of the tennis command in their conditioned variants.
The baseline methods also share the same task reward, the actor's observations of the task command, and the initial states.
Details of the baseline methods are provided in Appendix~\ref{app:baselines}.

As shown in \Cref{tab:cond}, the task-only policies fail to learn the strike, and adding an unconditional prior does not help.
In contrast, all three task-conditioned variants achieve a much higher strike rate and a lower landing error.
The unconditional variants fail because the hitting term of the task reward is nonzero only within about \SI{0.1}{\second} of the contact and near the interception point (Appendix~\ref{app:tasks_tennis}).
Exploration therefore rarely earns it, and the approach and backswing before the contact go unrewarded.
Among them, our proposed \method outperforms AMP and achieves task success similar to SMP. 
Overall, the results suggest that task conditioning guides the detailed movements that lead to sparse events, such as the stroke and the contact, by anticipating the swing.

\method improves motion quality in addition to task completion. \Cref{tab:cond} reports the Fréchet distance to the data as a measure of motion quality. 
Without any prior, the policy produces highly unnatural motions with the highest Fréchet distance. 
Adding a prior reduces it by an order of magnitude, and conditioning the prior on the task reduces it further for all three priors. 
This further indicates that the task condition helps the prior provide guidance toward natural motions that are relevant to the task, rather than toward the dataset as a whole.
AMP attains a lower Fréchet distance but also a lower strike rate.
This suggests that AMP focuses on reproducing the dataset rather than adapting to the task.
Among the priors, \method thus offers the best trade-off between task success and motion quality.
Details on the metrics and the evaluation protocol are given in Appendices~\ref{app:metrics} and~\ref{app:evaluation}.

We also evaluate the effect of task conditioning of our proposed \method in a denser reward setting with the box-carrying task.
Its task reward depends only on the state of the box and prescribes its course through lifting, transporting to the goal, and setting down (Appendix~\ref{app:tasks_box}).
We compare the conditional prior with the unconditional one and with the task reward alone (\Cref{tab:cond_box}), similar to tennis. 
The results show that the task reward alone achieves a place rate of $43\%$, whereas applying the unconditional prior increases it to $55\%$.
The conditional prior increases it further to $83\%$, showing that a task-informed prior guides the motion more directly toward completing the task.

\begin{table}[t]
    \caption{Motion priors on goal reaching and steering.}
    \label{tab:prior_comparison}
    \centering
    \footnotesize
    \begin{tabular}{lccc}
        \thickhline
        & & \multicolumn{2}{c}{Fréchet $\downarrow$} \\
        \cmidrule(lr){3-4}
        Variant & Reusable & Goal reaching & Steering \\
        \hline
        Task only & -- & $233 \pm 39$ & $193 \pm 14$ \\
        AMP & no & \tabbest{1.91 \pm 0.69} & \tabbest{2.24 \pm 0.12} \\
        AMP (frozen)$^{\dagger}$ & yes & $316 \pm 207$ & $253 \pm 115$ \\
        SMP & yes & $5.84 \pm 0.83$ & $4.11 \pm 0.19$ \\
        \method (ours) & yes & \tabsecond{5.42 \pm 0.50} & \tabsecond{4.04 \pm 0.16} \\
        \thickhline
    \end{tabular}
    \tabnote{Mean $\pm$ std over 3 seeds; Fréchet is relative to the dataset-to-dataset distance. Frozen priors are not trained with the policy; SMP and \method share theirs between the two tasks. Dark blue and bold: best mean per column; light blue: second best. $^{\dagger}$Discriminator of the other task's final AMP policy.}
\end{table}

\subsection{Reusable Prior with Explicit Reference}
\label{sec:exp_reuse}
\method obtains an explicit reference for the policy by projecting its motion through a pretrained flow-matching model.
This gives rise to two advantages. 
First, a single prior can be reused across multiple tasks.
Since the prior is trained once and stays frozen during policy learning, the style reward does not depend on the specific task setting. 
Second, the same prior can be used with different style reward compositions. 
The same projected reference can be restricted to selected parts of the body to make different style rewards.

\subsubsection{Reuse across tasks} 
\label{sec:exp_prior}
We evaluate whether a single prior, trained once and kept frozen, can guide policies toward natural motion across different tasks.
To this end, we train one \method prior on running data without task conditioning and use it for both goal reaching and steering.
We compare it with SMP~\citep{mu2025smp}, whose frozen prior is trained on the same data with the same network and likewise serves both tasks, and AMP, whose discriminator is trained together with the policy of each task and kept frozen for the other.
Task-only policies, trained without any style reward, serve as a baseline.
All policies achieve similar task completion. 
We report the Fréchet distance to held-out motion data under identical commands (\Cref{tab:prior_comparison}).

Without any style reward, policies produce motions far from the data, such as abrupt stops and turns.
All three priors bring the motion close to the data.
AMP yields the best motion quality, and \method and SMP show comparable results.
When the discriminator of AMP is instead kept frozen and reused for the other task, the Fréchet distance rises from $1.91$ to $316$ in goal reaching and from $2.24$ to $253$ in steering, even above the task-only policies.
In contrast, the frozen priors of \method and SMP keep the motion close to the data in both tasks.

\begin{table}[t]
    \caption{Style reward restricted to a part of the body.}
    \label{tab:partial_body}
    \centering
    \footnotesize
    \begin{tabular}{lccc}
        \thickhline
        & \multicolumn{3}{c}{Fréchet $\downarrow$} \\
        \cmidrule(lr){2-4}
        Style reward & Full pose & Lower body & Upper body \\
        \hline
        Full body & \tabbest{10.3 \pm 0.2} & \tabbest{6.3 \pm 0.1} & \tabbest{16.7 \pm 0.6} \\
        Lower body & $326 \pm 56$ & \tabsecond{9.0 \pm 2.7} & $735 \pm 131$ \\
        Upper body & \tabsecond{53.5 \pm 19.9} & $112 \pm 49$ & \tabsecond{17.5 \pm 0.5} \\
        \thickhline
    \end{tabular}
    \tabnote{Stylized goal reaching, averaged over the 14 styles; mean $\pm$ std over 3 seeds. Fréchet is relative to the dataset-to-dataset distance of each column. Dark blue and bold: best mean per column; light blue: second best.}
\end{table}

\begin{figure}[t]
    \centering
    \includegraphics[width=\columnwidth]{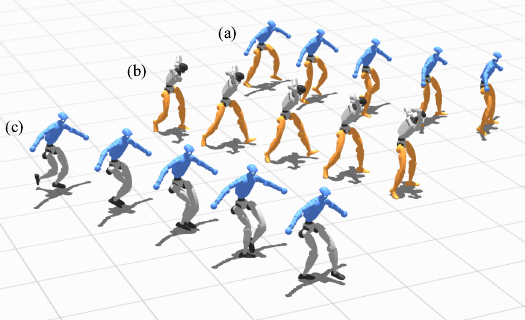}
    \caption{Stylized goal reaching in the style \emph{Star} with the style reward on (a) the full body, (b) the lower body, and (c) the upper body. All three start from the same state toward the same goal. Colored limbs are read by the style reward: legs in orange, waist and arms in blue.}
    \label{fig:partial_strip}
\end{figure}

\subsubsection{Reuse across body parts}
\label{sec:exp_partial}

The projection of \method provides an explicit reference, which allows the same prior to be reused across body parts.
In detail, the style reward tracks this reference as a product of per-group terms (\Cref{eq:style_reward}), so each part of the body can be included in or left out of the reward.

We test this on stylized locomotion with the style-conditioned walking prior by restricting the joint-position and key-body groups of the reward to the lower body (legs and feet) or to the upper body (waist, arms, torso, and wrists); the root groups are kept in both variants.
\Cref{tab:partial_body} reports the distance of the policy's poses to the data of the commanded style, separately for the two parts of the body.
Although the projection covers the whole body, each restricted reward keeps the part it reads close to the data, at a level similar to that of the full-body reward, while the part it does not include drifts far from the data toward whatever the task finds convenient.
All goals are reached in every variant, so the style reward on either part does not compete with the task.
\Cref{fig:partial_strip} shows a qualitative example of this restriction with three policies side by side in the style \emph{Star}, where the wide stance appears only when the reward reads the legs, and the spread arms only when it reads the arms.

\subsection{Ablation of the Projection Parameters}
\label{sec:exp_design}

The choice of projection parameters is central to \method, as it determines what the reference asks of the policy.
We therefore ablate the two projection parameters to examine how strongly the reference corrects the policy's motion and whether it follows the task condition.
The first is the projection time $\tproj$, which sets the amount of noise injected into the input.
The second is the number of integration steps $N$, which sets how finely the flow is integrated from the noised input.
Using the frozen tennis prior, we run three experiments.
We first examine offline how well perturbed motions are corrected (\Cref{sec:exp_design_quality}).
We then evaluate whether the reference follows a condition that contradicts the input (\Cref{sec:exp_design_cond}).
Finally, we examine whether these effects carry over to policy training (\Cref{sec:exp_design_online}).

\subsubsection{Perturbed motions}
\label{sec:exp_design_quality}
\begin{figure}[t]
    \centering
    \includegraphics[width=\columnwidth]{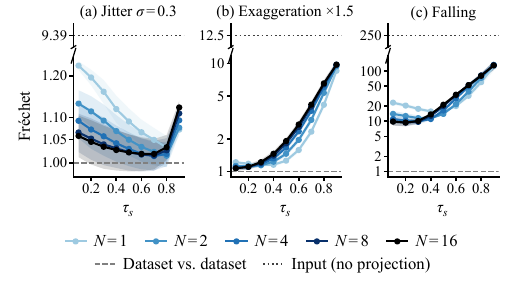}
    \caption{Fréchet distance of the references relative to the dataset-to-dataset distance, for jittery motion with $\sigma=0.3$, motion exaggerated by a factor of $1.5$, and falling motion under the tennis prior. The dotted line is the unprojected input. The y-axes are logarithmic; bands show the std over seeds.}
    \label{fig:design_frechet}
\end{figure}

During exploration, the policy produces motions that deviate from the data, yet these motions usually still carry a meaningful structure.
The projection should therefore move them toward the data without discarding this structure.
Too little noise leaves the deviation uncorrected, whereas too much noise discards the motion and lets the prior regenerate it, which can drift from the data when the input was already close to it.
To find where this balance lies, we project jittery motion with $\sigma = 0.3$, motion exaggerated by a factor of $1.5$, and falling motion with $\tproj$ from $0.1$ to $0.9$ and $N \in \{1, 2, 4, 8, 16\}$, and measure the Fréchet distance of the references to the data (\Cref{fig:design_frechet}).

Jitter is removed at almost any $\tproj$, and the reference is closest to the data around $\tproj = 0.6$ to $0.7$, where the injected noise covers the jitter while the underlying natural motion is preserved.
Exaggerated and falling motions, in contrast, deviate coherently over the whole window and are corrected more at smaller $\tproj$, as more of the deviation is covered by noise.
The number of integration steps amplifies the effect of $\tproj$.
At small $\tproj$, most of the input is covered by noise, so more steps bring the reference closer to the data.
At large $\tproj$, more of the input remains after noising, so more steps instead reproduce it more faithfully.

\subsubsection{Mismatched conditions}
\label{sec:exp_design_cond}
With a conditional prior, the projection is pulled in two directions.
The reference can preserve the current motion of the policy, or it can change it to fit the condition.
The two conflict when the policy plays a stroke other than the commanded one, as often happens early in training.
To see which one the reference follows, we give the projection a condition that disagrees with the input.
We project forehand windows with a backhand condition and vice versa, and we project falling motion, which has no stroke, with either condition.

\begin{figure}[t]
    \centering
    \includegraphics[width=\columnwidth]{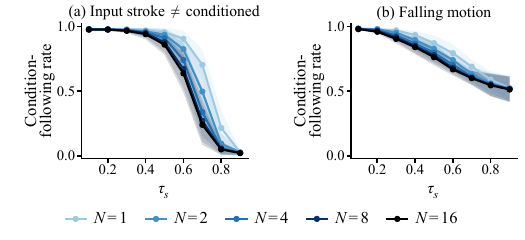}
    \caption{Effect of the projection time $\tproj$ and the number of integration steps $N$ on the stroke of the reference under the frozen tennis prior. (a) Dataset windows of one stroke projected with the hit-target condition of the other stroke. (b) Windows of falling motion projected with a forehand or backhand condition. Curves and bands show the fraction of references classified as the conditioned stroke over three seeds.}
    \label{fig:design_coverage}
\end{figure}

Each reference is classified as a forehand or a backhand by a nearest-neighbor vote against the labeled data, and \Cref{fig:design_coverage} reports the fraction classified as the conditioned stroke.
The projection time $\tproj$ acts as a dial between the condition and the input.
At small $\tproj$, the reference shows the conditioned stroke even when the input is the other stroke, whereas at large $\tproj$, it keeps the stroke of the input, and references of falling motion drop to chance.
More integration steps shift this transition to smaller $\tproj$: for a mismatched input at $\tproj = 0.7$, the reference shows the conditioned stroke in $70\%$ of the cases with one step and in $24\%$ with sixteen.

\subsubsection{Trained policies}
\label{sec:exp_design_online}
We now train policies with different projection parameters.
Unlike in the offline experiments, the policy adapts to the style reward during training, so its motion and the reference change together.
We examine whether the offline trends still hold for the motion of the trained policies and whether the policies learn the commanded strokes.

\begin{figure}[t]
    \centering
    \includegraphics[width=\columnwidth]{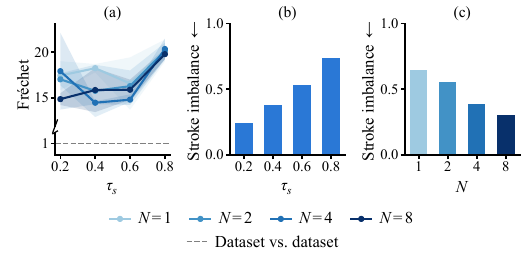}
    \caption{Effect of the projection time $\tproj$ and the number of integration steps $N$ on motion quality and mode preservation of policies trained with the tennis prior. (a) Fréchet distance of the policy's windows \SIrange{0.4}{0.6}{\second} before contact, relative to the dataset-to-dataset distance; mean $\pm$ std over three seeds. (b, c) Stroke imbalance, averaged over the twelve policies that share a value of $\tproj$ (b) or of $N$ (c).}
    \label{fig:design_online}
\end{figure}

We train tennis policies on a $4 \times 4$ grid of $\tproj \in \{0.2, 0.4, 0.6, 0.8\}$ and $N \in \{1, 2, 4, 8\}$ with three seeds each.
Every policy is evaluated on an even split of forehand and backhand commands (\Cref{fig:design_online}).
We measure the Fréchet distance of the policy's motion and the stroke imbalance $|p_{\mathrm{FH}} - p_{\mathrm{BH}}| / (p_{\mathrm{FH}} + p_{\mathrm{BH}})$.
Here, $p_{\mathrm{FH}}$ and $p_{\mathrm{BH}}$ are the fractions of forehand and backhand commands answered with a clean hit of the commanded stroke.
An imbalance of 0 means that both strokes are played equally well, and a value near 1 means that the policy has collapsed to one of them.

As in the offline experiments, the motion is farthest from the data at the largest $\tproj$, the smallest $\tproj$ is also slightly worse, and the number of steps has little effect (\Cref{fig:design_online}a).
The projection time also determines whether the policy follows the command.
The imbalance grows from $0.24$ at $\tproj = 0.2$ to $0.73$ at $\tproj = 0.8$ (\Cref{fig:design_online}b).
A small $\tproj$ corrects the policy toward the commanded stroke, whereas a large $\tproj$ keeps the policy's own swing, and most policies settle on the forehand.
This matches the offline result that a large $\tproj$ follows the input rather than the condition.
More integration steps reduce the imbalance from $0.65$ with one step to $0.30$ with eight (\Cref{fig:design_online}c).
This differs from the offline result, where more steps favor the input, and we attribute it to the interplay between the task and style rewards during training.

In summary, the projection time is the main parameter and is chosen per task: it sets how far the policy's motion may deviate from the data and from the task condition before the reward corrects it.
More integration steps integrate the flow more accurately, but each step is one more evaluation of the prior per control step, so we use $N = 4$ for all tasks.

\subsection{Deployment}
\label{sec:exp_deploy}

We deploy the goal-reaching, steering, stylized-locomotion, box-carrying, and tennis policies on the Unitree G1 without fine-tuning.
The actor observes a history of proprioception and the command, which the onboard IMU and joint encoders provide. 
No estimate of the base linear velocity is needed.
An external motion-capture system is used to track the global poses of the robot and of the objects (the box and the tennis ball).

In tennis, we estimate the velocity of the ball from its tracked positions and roll out the flight model of Appendix~\ref{app:flight_model} from this state.
The predicted trajectory yields the interception point, the time to contact, and the velocity with which the ball has to leave the racket to reach the landing target.
Together with the positions of the ball and of the landing target, they form the command of the policy and are updated until shortly before the contact.

\Cref{fig:hardware} shows the robot on the four tasks.
The robot comes to a standstill at the goal (a), holds the posture of each style while it steps and changes it within a step when the command switches, all with one policy (b), crouches to the box and lifts it with both hands (c), and prepares each stroke and follows through, with footwork between the balls (d)--(e).

\section{Conclusion}
\label{sec:conclusion}

Rewarding the outcome of a whole-body skill does not determine the motion that achieves it.
We presented \method, a flow-matching motion prior that fills this gap by steering the policy toward motion that is close to human demonstrations and fits the task at each moment.
At its core is a projection: at every control step, the policy's recent motion is partially noised and integrated through a frozen flow model, optionally conditioned on the task context, and the projected motion serves as the reference of a tracking-style reward.
Our experiments show that the projection corrects a motion toward the data while preserving its overall structure, and that the resulting reward keeps the policies close to the motion data, on par with a score-matching prior.
The task condition makes a timed skill such as the tennis stroke learnable where the task reward alone is not sufficient.
One frozen prior serves several tasks, and because the projection yields an explicit motion for the policy to track, the reward can be inspected and restricted to parts of the body.
All policies transfer to the Unitree G1 without fine-tuning.
Through the projection, a frozen generative model can thus serve both as a description of how to move and as context-dependent guidance for what a task requires.

\method also comes with limitations that point to future work.
The projection time has to be tuned per task, since it balances how much the reference keeps of the policy's own motion against how strongly it follows the prior and its condition (\Cref{sec:exp_design}).
Our ablation guides this choice.
A small projection time makes the reference follow the condition, as needed in tennis, whereas a large one preserves motions that are already close to the data.
Moreover, since the reference is projected from the policy's own motion, the style reward tends to reinforce the motion that the policy already plays.
Even when the prior covers several modes, the policy may therefore learn only one of them, as some tennis policies settle on the forehand and rarely play the commanded backhand.
Encouraging the policy to explore and retain the modes of the prior is a direction for future work.

\section*{Acknowledgments}
We thank Sitong Chen and Jaeryeong Kim for their help with the hardware experiments.

\bibliographystyle{IEEEtranN}
\bibliography{root.bib}

@IEEEtranBSTCTL{IEEEexample:BSTcontrol,
  CTLuse_forced_etal       = "yes",
  CTLmax_names_forced_etal = "6",
  CTLnames_show_etal       = "6",
  CTLdash_repeated_names   = "no",
  CTLuse_url               = "no"
}

@inproceedings{lipman2023flow,
  title={Flow Matching for Generative Modeling},
  author={Lipman, Yaron and Chen, Ricky T. Q. and Ben-Hamu, Heli and Nickel, Maximilian and Le, Matt},
  booktitle={International Conference on Learning Representations (ICLR)},
  year={2023}
}

@inproceedings{liu2023flow,
  title={Flow Straight and Fast: Learning to Generate and Transfer Data with Rectified Flow},
  author={Liu, Xingchao and Gong, Chengyue and Liu, Qiang},
  booktitle={International Conference on Learning Representations (ICLR)},
  year={2023}
}

@inproceedings{meng2022sdedit,
  title={{SDEdit}: Guided Image Synthesis and Editing with Stochastic Differential Equations},
  author={Meng, Chenlin and He, Yutong and Song, Yang and Song, Jiaming and Wu, Jiajun and Zhu, Jun-Yan and Ermon, Stefano},
  booktitle={International Conference on Learning Representations (ICLR)},
  year={2022}
}

@article{peng2018deepmimic,
  title={{DeepMimic}: Example-Guided Deep Reinforcement Learning of Physics-Based Character Skills},
  author={Peng, Xue Bin and Abbeel, Pieter and Levine, Sergey and van de Panne, Michiel},
  journal={ACM Transactions on Graphics},
  volume={37},
  number={4},
  year={2018}
}

@article{mu2025smp,
  title={{SMP}: Reusable Score-Matching Motion Priors for Physics-Based Character Control},
  author={Mu, Yuxuan and Zhang, Ziyu and Shi, Yi and Yang, Dun and Matsumoto, Minami and Imamura, Kotaro and Tevet, Guy and Guo, Chuan and Taylor, Michael and Shu, Chang and Xi, Pengcheng and Peng, Xue Bin},
  journal={ACM Transactions on Graphics (SIGGRAPH)},
  year={2026}
}

@article{schulman2017ppo,
  title={Proximal Policy Optimization Algorithms},
  author={Schulman, John and Wolski, Filip and Dhariwal, Prafulla and Radford, Alec and Klimov, Oleg},
  journal={arXiv preprint arXiv:1707.06347},
  year={2017}
}

@inproceedings{schulman2016gae,
  title={High-Dimensional Continuous Control Using Generalized Advantage Estimation},
  author={Schulman, John and Moritz, Philipp and Levine, Sergey and Jordan, Michael and Abbeel, Pieter},
  booktitle={International Conference on Learning Representations (ICLR)},
  year={2016}
}

@inproceedings{peebles2023dit,
  title={Scalable Diffusion Models with Transformers},
  author={Peebles, William and Xie, Saining},
  booktitle={IEEE/CVF International Conference on Computer Vision (ICCV)},
  year={2023}
}

@inproceedings{chen2024pixart,
  title={{PixArt}-$\alpha$: Fast Training of Diffusion Transformer for Photorealistic Text-to-Image Synthesis},
  author={Chen, Junsong and Yu, Jincheng and Ge, Chongjian and Yao, Lewei and Xie, Enze and Wu, Yue and Wang, Zhongdao and Kwok, James and Luo, Ping and Lu, Huchuan and Li, Zhenguo},
  booktitle={International Conference on Learning Representations (ICLR)},
  year={2024}
}

@inproceedings{zhou2019continuity,
  title={On the Continuity of Rotation Representations in Neural Networks},
  author={Zhou, Yi and Barnes, Connelly and Lu, Jingwan and Yang, Jimei and Li, Hao},
  booktitle={IEEE/CVF Conference on Computer Vision and Pattern Recognition (CVPR)},
  year={2019}
}

@inproceedings{mysore2022multicritic,
  title={Multi-Critic Actor Learning: Teaching {RL} Policies to Act with Style},
  author={Mysore, Siddharth and Cheng, George and Zhao, Yunqi and Saenko, Kate and Wu, Meng},
  booktitle={International Conference on Learning Representations (ICLR)},
  year={2022}
}

@inproceedings{zargarbashi2024robotkeyframing,
  title={{RobotKeyframing}: Learning Locomotion with High-Level Objectives via Mixture of Dense and Sparse Rewards},
  author={Zargarbashi, Fatemeh and Cheng, Jin and Kang, Dongho and Sumner, Robert and Coros, Stelian},
  booktitle={Conference on Robot Learning (CoRL)},
  year={2024}
}

@article{harvey2020lafan,
  title={Robust Motion In-betweening},
  author={Harvey, F{\'e}lix G. and Yurick, Mike and Nowrouzezahrai, Derek and Pal, Christopher},
  journal={ACM Transactions on Graphics},
  volume={39},
  number={4},
  year={2020}
}

@article{mason2022style,
  title={Real-Time Style Modelling of Human Locomotion via Feature-Wise Transformations and Local Motion Phases},
  author={Mason, Ian and Starke, Sebastian and Komura, Taku},
  journal={Proceedings of the ACM on Computer Graphics and Interactive Techniques},
  volume={5},
  number={1},
  year={2022}
}

@inproceedings{yang2025omniretarget,
  title={{OmniRetarget}: Interaction-Preserving Data Generation for Humanoid Whole-Body Loco-Manipulation and Scene Interaction},
  author={Yang, Lujie and Huang, Xiaoyu and Wu, Zhen and Kanazawa, Angjoo and Abbeel, Pieter and Sferrazza, Carmelo and Liu, C. Karen and Duan, Rocky and Shi, Guanya},
  booktitle={IEEE International Conference on Robotics and Automation (ICRA)},
  year={2026}
}

@article{zakka2026mjlab,
  title={mjlab: A Lightweight Framework for {GPU}-Accelerated Robot Learning},
  author={Zakka, Kevin and Liao, Qiayuan and Yi, Brent and Le Lay, Louis and Sreenath, Koushil and Abbeel, Pieter},
  journal={arXiv preprint arXiv:2601.22074},
  year={2026}
}

@inproceedings{shen2024gvhmr,
  title={World-Grounded Human Motion Recovery via Gravity-View Coordinates},
  author={Shen, Zehong and Pi, Huaijin and Xia, Yan and Cen, Zhi and Peng, Sida and Hu, Zechen and Bao, Hujun and Hu, Ruizhen and Zhou, Xiaowei},
  booktitle={SIGGRAPH Asia Conference Papers},
  year={2024}
}

@inproceedings{araujo2025gmr,
  title={Retargeting Matters: General Motion Retargeting for Humanoid Motion Tracking},
  author={Araujo, Joao Pedro and Ze, Yanjie and Xu, Pei and Wu, Jiajun and Liu, C. Karen},
  booktitle={IEEE International Conference on Robotics and Automation (ICRA)},
  year={2026}
}

@article{li2023omomo,
  title={Object Motion Guided Human Motion Synthesis},
  author={Li, Jiaman and Wu, Jiajun and Liu, C. Karen},
  journal={ACM Transactions on Graphics},
  volume={42},
  number={6},
  year={2023}
}

@article{demler2026caltennis,
  title={{CalTennis}: Large Multi-View Tennis Video Dataset and Benchmark of Monocular-to-{3D} Pose Estimation},
  author={Demler, Ilona and Xie, Xinran and Werner, Blake and Szczuka, Anna and Perona, Pietro},
  journal={arXiv preprint arXiv:2606.20542},
  year={2026}
}

@misc{lafan1retargeted2025,
  title={{LAFAN1} Retargeting Dataset},
  author={{lvhaidong}},
  howpublished={\url{https://huggingface.co/datasets/lvhaidong/LAFAN1_Retargeting_Dataset}},
  note={{Hugging Face} dataset},
  year={2025}
}

@article{peng2021amp,
  title={{AMP}: Adversarial Motion Priors for Stylized Physics-Based Character Control},
  author={Peng, Xue Bin and Ma, Ze and Abbeel, Pieter and Levine, Sergey and Kanazawa, Angjoo},
  journal={ACM Transactions on Graphics},
  volume={40},
  number={4},
  year={2021}
}

@article{peng2022ase,
  title={{ASE}: Large-Scale Reusable Adversarial Skill Embeddings for Physically Simulated Characters},
  author={Peng, Xue Bin and Guo, Yunrong and Halper, Lina and Levine, Sergey and Fidler, Sanja},
  journal={ACM Transactions on Graphics},
  volume={41},
  number={4},
  year={2022}
}

@inproceedings{tessler2023calm,
  title={{CALM}: Conditional Adversarial Latent Models for Directable Virtual Characters},
  author={Tessler, Chen and Kasten, Yoni and Guo, Yunrong and Mannor, Shie and Chechik, Gal and Peng, Xue Bin},
  booktitle={ACM SIGGRAPH Conference Proceedings},
  year={2023}
}

@article{liao2025beyondmimic,
  title={{BeyondMimic}: From Motion Tracking to Versatile Humanoid Control via Guided Diffusion},
  author={Liao, Qiayuan and Truong, Takara E. and Huang, Xiaoyu and Gao, Yuman and Tevet, Guy and Sreenath, Koushil and Liu, C. Karen},
  journal={arXiv preprint arXiv:2508.08241},
  year={2025}
}

@inproceedings{liao2026supermimic,
  title={{SuperMimic}: Scaling Humanoid Whole-Body Control with One Differentiable World Model},
  author={Liao, Qiayuan and Shao, Yiyang and Long, Junfeng and Zakka, Kevin and Horowitz, Roberto and Abbeel, Pieter and Sreenath, Koushil},
  booktitle={Conference on Robot Learning (CoRL)},
  year={2026}
}

@article{weng2025hdmi,
  title={{HDMI}: Learning Interactive Humanoid Whole-Body Control from Human Videos},
  author={Weng, Haoyang and Li, Yitang and Sobanbabu, Nikhil and Wang, Zihan and Luo, Zhengyi and He, Tairan and Ramanan, Deva and Shi, Guanya},
  journal={arXiv preprint arXiv:2509.16757},
  year={2025}
}

@article{zhao2025resmimic,
  title={{ResMimic}: From General Motion Tracking to Humanoid Whole-Body Loco-Manipulation via Residual Learning},
  author={Zhao, Siheng and Ze, Yanjie and Wang, Yue and Liu, C. Karen and Abbeel, Pieter and Shi, Guanya and Duan, Rocky},
  journal={arXiv preprint arXiv:2510.05070},
  year={2025}
}

@article{luo2025sonic,
  title={{SONIC}: Supersizing Motion Tracking for Natural Humanoid Whole-Body Control},
  author={Luo, Zhengyi and Yuan, Ye and Wang, Tingwu and Li, Chenran and Casta{\~n}eda, Fernando and Chen, Sirui and Cao, Zi-Ang and Li, Jiefeng and Minor, David and Ben, Qingwei and Park, Jinhyung and Sami, David and Wang, Zi and Da, Xingye and Ding, Runyu and Hogg, Cyrus and Song, Lina and Lim, Edy and Jeong, Eugene and He, Tairan and Xue, Haoru and Xiao, Wenli and Yuen, Simon and Kautz, Jan and Chang, Yan and Iqbal, Umar and Fan, Linxi and Zhu, Yuke},
  journal={Science Robotics},
  volume={11},
  number={117},
  pages={eaed4592},
  year={2026}
}

@inproceedings{luo2023phc,
  title={Perpetual Humanoid Control for Real-time Simulated Avatars},
  author={Luo, Zhengyi and Cao, Jinkun and Winkler, Alexander and Kitani, Kris and Xu, Weipeng},
  booktitle={IEEE/CVF International Conference on Computer Vision (ICCV)},
  year={2023}
}

@inproceedings{cheng2024exbody,
  title={Expressive Whole-Body Control for Humanoid Robots},
  author={Cheng, Xuxin and Ji, Yandong and Chen, Junming and Yang, Ruihan and Yang, Ge and Wang, Xiaolong},
  booktitle={Robotics: Science and Systems (RSS)},
  year={2024}
}

@inproceedings{he2024h2o,
  title={Learning Human-to-Humanoid Real-Time Whole-Body Teleoperation},
  author={He, Tairan and Luo, Zhengyi and Xiao, Wenli and Zhang, Chong and Kitani, Kris and Liu, Changliu and Shi, Guanya},
  booktitle={IEEE/RSJ International Conference on Intelligent Robots and Systems (IROS)},
  year={2024}
}

@inproceedings{he2024omnih2o,
  title={{OmniH2O}: Universal and Dexterous Human-to-Humanoid Whole-Body Teleoperation and Learning},
  author={He, Tairan and Luo, Zhengyi and He, Xialin and Xiao, Wenli and Zhang, Chong and Zhang, Weinan and Kitani, Kris and Liu, Changliu and Shi, Guanya},
  booktitle={Conference on Robot Learning (CoRL)},
  year={2024}
}

@inproceedings{fu2024humanplus,
  title={{HumanPlus}: Humanoid Shadowing and Imitation from Humans},
  author={Fu, Zipeng and Zhao, Qingqing and Wu, Qi and Wetzstein, Gordon and Finn, Chelsea},
  booktitle={Conference on Robot Learning (CoRL)},
  year={2024}
}

@inproceedings{he2025asap,
  title={{ASAP}: Aligning Simulation and Real-World Physics for Learning Agile Humanoid Whole-Body Skills},
  author={He, Tairan and Gao, Jiawei and Xiao, Wenli and Zhang, Yuanhang and Wang, Zi and Wang, Jiashun and Luo, Zhengyi and He, Guanqi and Sobanbabu, Nikhil and Pan, Chaoyi and Yi, Zeji and Qu, Guannan and Kitani, Kris and Hodgins, Jessica and Fan, Linxi and Zhu, Yuke and Liu, Changliu and Shi, Guanya},
  booktitle={Robotics: Science and Systems (RSS)},
  year={2025}
}

@article{chen2025gmt,
  title={{GMT}: General Motion Tracking for Humanoid Whole-Body Control},
  author={Chen, Zixuan and Ji, Mazeyu and Cheng, Xuxin and Peng, Xuanbin and Peng, Xue Bin and Wang, Xiaolong},
  journal={arXiv preprint arXiv:2506.14770},
  year={2025}
}

@inproceedings{ze2025twist,
  title={{TWIST}: Teleoperated Whole-Body Imitation System},
  author={Ze, Yanjie and Chen, Zixuan and Ara{\'u}jo, Jo{\~a}o Pedro and Cao, Zi-ang and Peng, Xue Bin and Wu, Jiajun and Liu, C. Karen},
  booktitle={Conference on Robot Learning (CoRL)},
  year={2025}
}

@inproceedings{allshire2025videomimic,
  title={Visual Imitation Enables Contextual Humanoid Control},
  author={Allshire, Arthur and Choi, Hongsuk and Zhang, Junyi and McAllister, David and Zhang, Anthony and Kim, Chung Min and Darrell, Trevor and Abbeel, Pieter and Malik, Jitendra and Kanazawa, Angjoo},
  booktitle={Conference on Robot Learning (CoRL)},
  year={2025}
}

@inproceedings{dao2024box,
  title={Sim-to-Real Learning for Humanoid Box Loco-Manipulation},
  author={Dao, Jeremy and Duan, Helei and Fern, Alan},
  booktitle={IEEE International Conference on Robotics and Automation (ICRA)},
  year={2024}
}

@inproceedings{tevet2023mdm,
  title={Human Motion Diffusion Model},
  author={Tevet, Guy and Raab, Sigal and Gordon, Brian and Shafir, Yonatan and Cohen-Or, Daniel and Bermano, Amit H.},
  booktitle={International Conference on Learning Representations (ICLR)},
  year={2023}
}

@inproceedings{yuan2023physdiff,
  title={{PhysDiff}: Physics-Guided Human Motion Diffusion Model},
  author={Yuan, Ye and Song, Jiaming and Iqbal, Umar and Vahdat, Arash and Kautz, Jan},
  booktitle={IEEE/CVF International Conference on Computer Vision (ICCV)},
  year={2023}
}

@inproceedings{tevet2025closd,
  title={{CLoSD}: Closing the Loop between Simulation and Diffusion for Multi-Task Character Control},
  author={Tevet, Guy and Raab, Sigal and Cohan, Setareh and Reda, Daniele and Luo, Zhengyi and Peng, Xue Bin and Bermano, Amit H. and van de Panne, Michiel},
  booktitle={International Conference on Learning Representations (ICLR)},
  year={2025}
}

@article{xie2026textop,
  title={{TextOp}: Real-time Interactive Text-Driven Humanoid Robot Motion Generation and Control},
  author={Xie, Weiji and Zheng, Jiakun and Han, Jinrui and Shi, Jiyuan and Zhang, Weinan and Bai, Chenjia and Li, Xuelong},
  journal={arXiv preprint arXiv:2602.07439},
  year={2026}
}

@inproceedings{luo2024pulse,
  title={Universal Humanoid Motion Representations for Physics-Based Control},
  author={Luo, Zhengyi and Cao, Jinkun and Merel, Josh and Winkler, Alexander and Huang, Jing and Kitani, Kris and Xu, Weipeng},
  booktitle={International Conference on Learning Representations (ICLR)},
  year={2024}
}

@article{tessler2024maskedmimic,
  title={{MaskedMimic}: Unified Physics-Based Character Control Through Masked Motion Inpainting},
  author={Tessler, Chen and Guo, Yunrong and Nabati, Ofir and Chechik, Gal and Peng, Xue Bin},
  journal={ACM Transactions on Graphics},
  volume={43},
  number={6},
  year={2024}
}

@inproceedings{truong2024pdp,
  title={{PDP}: Physics-Based Character Animation via Diffusion Policy},
  author={Truong, Takara E. and Piseno, Michael and Xie, Zhaoming and Liu, C. Karen},
  booktitle={SIGGRAPH Asia Conference Papers},
  year={2024}
}

@article{huang2025diffusecloc,
  title={{Diffuse-CLoC}: Guided Diffusion for Physics-based Character Look-ahead Control},
  author={Huang, Xiaoyu and Truong, Takara E. and Zhang, Yunbo and Yu, Fangzhou and Sleiman, Jean-Pierre and Hodgins, Jessica and Sreenath, Koushil and Farshidian, Farbod},
  journal={ACM Transactions on Graphics},
  volume={44},
  number={4},
  year={2025}
}

@inproceedings{escontrela2022amp,
  title={Adversarial Motion Priors Make Good Substitutes for Complex Reward Functions},
  author={Escontrela, Alejandro and Peng, Xue Bin and Yu, Wenhao and Zhang, Tingnan and Iscen, Atil and Goldberg, Ken and Abbeel, Pieter},
  booktitle={IEEE/RSJ International Conference on Intelligent Robots and Systems (IROS)},
  year={2022}
}

@inproceedings{vollenweider2023multiamp,
  title={Advanced Skills through Multiple Adversarial Motion Priors in Reinforcement Learning},
  author={Vollenweider, Eric and Bjelonic, Marko and Klemm, Victor and Rudin, Nikita and Lee, Joonho and Hutter, Marco},
  booktitle={IEEE International Conference on Robotics and Automation (ICRA)},
  year={2023}
}

@inproceedings{zhang2024adam,
  title={Whole-body Humanoid Robot Locomotion with Human Reference},
  author={Zhang, Qiang and Cui, Peter and Yan, David and Sun, Jingkai and Duan, Yiqun and Han, Gang and Zhao, Wen and Zhang, Weining and Guo, Yijie and Zhang, Arthur and Xu, Renjing},
  booktitle={IEEE/RSJ International Conference on Intelligent Robots and Systems (IROS)},
  pages={11225--11231},
  year={2024}
}

@inproceedings{alvarez2025costume,
  title={Learning to Walk in Costume: Adversarial Motion Priors for Aesthetically Constrained Humanoids},
  author={Alvarez, Arturo Flores and Zargarbashi, Fatemeh and Liu, Havel and Wang, Shiqi and Edwards, Liam and Anz, Jessica and Xu, Alex and Shi, Fan and Coros, Stelian and Hong, Dennis W.},
  booktitle={IEEE-RAS International Conference on Humanoid Robots (Humanoids)},
  year={2025}
}

@inproceedings{hassan2023interphys,
  title={Synthesizing Physical Character-Scene Interactions},
  author={Hassan, Mohamed and Guo, Yunrong and Wang, Tingwu and Black, Michael J. and Fidler, Sanja and Peng, Xue Bin},
  booktitle={ACM SIGGRAPH Conference Proceedings},
  year={2023}
}

@inproceedings{bae2023pmp,
  title={{PMP}: Learning to Physically Interact with Environments using Part-wise Motion Priors},
  author={Bae, Jinseok and Won, Jungdam and Lim, Donggeun and Min, Cheol-Hui and Kim, Young Min},
  booktitle={ACM SIGGRAPH Conference Proceedings},
  year={2023}
}

@inproceedings{gao2024coohoi,
  title={{CooHOI}: Learning Cooperative Human-Object Interaction with Manipulated Object Dynamics},
  author={Gao, Jiawei and Wang, Ziqin and Xiao, Zeqi and Wang, Jingbo and Wang, Tai and Cao, Jinkun and Hu, Xiaolin and Liu, Si and Dai, Jifeng and Pang, Jiangmiao},
  booktitle={Advances in Neural Information Processing Systems (NeurIPS)},
  year={2024}
}

@inproceedings{diwan2025near,
  title={Noise-conditioned Energy-based Annealed Rewards ({NEAR}): A Generative Framework for Imitation Learning from Observation},
  author={Diwan, Anish Abhijit and Urain, Julen and Kober, Jens and Peters, Jan},
  booktitle={International Conference on Learning Representations (ICLR)},
  year={2025}
}

@article{chen2026diffcast,
  title={Constraint-Aware Diffusion Priors for High-Fidelity and Versatile Quadruped Locomotion},
  author={Chen, Jianhui and Zhan, Ruixin and Liu, Liu and Cai, Yang and Li, Ziqiao},
  journal={arXiv preprint arXiv:2605.08804},
  year={2026}
}

@article{liu2022football,
  title={From Motor Control to Team Play in Simulated Humanoid Football},
  author={Liu, Siqi and Lever, Guy and Wang, Zhe and Merel, Josh and Eslami, S. M. Ali and Hennes, Daniel and Czarnecki, Wojciech M. and Tassa, Yuval and Omidshafiei, Shayegan and Abdolmaleki, Abbas and Siegel, Noah Y. and Hasenclever, Leonard and Marris, Luke and Tunyasuvunakool, Saran and Song, H. Francis and Wulfmeier, Markus and Muller, Paul and Haarnoja, Tuomas and Tracey, Brendan D. and Tuyls, Karl and Graepel, Thore and Heess, Nicolas},
  journal={Science Robotics},
  volume={7},
  number={69},
  year={2022}
}

@article{zhang2023vid2player3d,
  title={Learning Physically Simulated Tennis Skills from Broadcast Videos},
  author={Zhang, Haotian and Yuan, Ye and Makoviychuk, Viktor and Guo, Yunrong and Fidler, Sanja and Peng, Xue Bin and Fatahalian, Kayvon},
  journal={ACM Transactions on Graphics},
  volume={42},
  number={4},
  year={2023}
}

@inproceedings{wang2024tabletennis,
  title={Strategy and Skill Learning for Physics-based Table Tennis Animation},
  author={Wang, Jiashun and Hodgins, Jessica and Won, Jungdam},
  booktitle={ACM SIGGRAPH Conference Papers},
  year={2024}
}

@article{haarnoja2024soccer,
  title={Learning Agile Soccer Skills for a Bipedal Robot with Deep Reinforcement Learning},
  author={Haarnoja, Tuomas and Moran, Ben and Lever, Guy and Huang, Sandy H. and Tirumala, Dhruva and Humplik, Jan and Wulfmeier, Markus and Tunyasuvunakool, Saran and Siegel, Noah Y. and Hafner, Roland and Bloesch, Michael and Hartikainen, Kristian and Byravan, Arunkumar and Hasenclever, Leonard and Tassa, Yuval and Sadeghi, Fereshteh and Batchelor, Nathan and Casarini, Federico and Saliceti, Stefano and Game, Charles and Sreendra, Neil and Patel, Kushal and Gwira, Marlon and Huber, Andrea and Hurley, Nicole and Nori, Francesco and Hadsell, Raia and Heess, Nicolas},
  journal={Science Robotics},
  volume={9},
  number={89},
  year={2024}
}

@article{wang2025soccer,
  title={Learning Vision-Driven Reactive Soccer Skills for Humanoid Robots},
  author={Wang, Yushi and Luo, Changsheng and Chen, Penghui and Liu, Jianran and Sun, Weijian and Guo, Tong and Yang, Kechang and Hu, Biao and Zhang, Yangang and Zhao, Mingguo},
  journal={Science Robotics},
  volume={11},
  pages={eaed1152},
  year={2026}
}

@article{ren2025goalkeeper,
  title={Humanoid Goalkeeper: Learning from Position Conditioned Task-Motion Constraints},
  author={Ren, Junli and Long, Junfeng and Huang, Tao and Wang, Huayi and Wang, Zirui and Jia, Feiyu and Zhang, Wentao and Wang, Jingbo and Luo, Ping and Pang, Jiangmiao},
  journal={IEEE Robotics and Automation Letters (RA-L)},
  year={2026}
}

@article{su2025hitter,
  title={{HITTER}: A {HumanoId Table TEnnis} Robot via Hierarchical Planning and Learning},
  author={Su, Zhi and Zhang, Bike and Rahmanian, Nima and Gao, Yuman and Liao, Qiayuan and Regan, Caitlin and Sreenath, Koushil and Sastry, S. Shankar},
  journal={arXiv preprint arXiv:2508.21043},
  year={2025}
}

@inproceedings{hu2025tabletennis,
  title={{PACE}: Physics Augmentation for Coordinated End-to-end Reinforcement Learning toward Versatile Humanoid Table Tennis},
  author={Hu, Muqun and Chen, Wenxi and Li, Wenjing and Mandali, Falak and He, Zijian and Zhang, Renhong and Krisna, Praveen and Christian, Katherine and Benaharon, Leo and Ma, Dizhi and Ramani, Karthik and Gu, Yan},
  booktitle={IEEE International Conference on Robotics and Automation (ICRA)},
  year={2026}
}

@article{ma2025badminton,
  title={Learning Coordinated Badminton Skills for Legged Manipulators},
  author={Ma, Yuntao and Cramariuc, Andrei and Farshidian, Farbod and Hutter, Marco},
  journal={Science Robotics},
  volume={10},
  number={102},
  year={2025}
}

@article{liu2025badminton,
  title={Humanoid Whole-Body Badminton via an Annealed Reinforcement Learning Curriculum},
  author={Liu, Chenhao and Jiang, Leyun and Tian, Ningyuan and Wang, Yibo and Yao, Kairan and Fu, Jinchen and Ren, Xiaoyu},
  journal={arXiv preprint arXiv:2511.11218},
  year={2025}
}

@article{chen2026badminton,
  title={Learning Human-Like Badminton Skills for Humanoid Robots},
  author={Chen, Yeke and Dong, Shihao and Ji, Xiaoyu and Sun, Jingkai and Luo, Zeren and Zhao, Liu and Zhang, Jiahui and Li, Wanyue and Ma, Ji and Xu, Bowen and Han, Yimin and Li, Xuanyi and Zhao, Yudong and Li, Liyun and Lu, Peng},
  journal={arXiv preprint arXiv:2602.08370},
  year={2026}
}

@article{zhang2026latent,
  title={Learning Athletic Humanoid Tennis Skills from Imperfect Human Motion Data},
  author={Zhang, Zhikai and Lu, Haofei and Lian, Yunrui and Chen, Ziqing and Liu, Yun and Lin, Chenghuai and Xue, Han and Zeng, Zicheng and Qi, Zekun and Zheng, Shaolin and Luan, Qing and Wang, Jingbo and Xing, Junliang and Wang, He and Yi, Li},
  journal={arXiv preprint arXiv:2603.12686},
  year={2026}
}

@inproceedings{huang2026adapt,
  title={Towards Professional Tennis Styles for Humanoid Robots with Adaptive Motion Planning and Tracking},
  author={Huang, Tao and Liu, Ruofei and Tang, Xuchen and Zhang, Xinyin and Ren, Junli and Wang, Huayi and Jia, Feiyu and Qi, Yukai and Yin, Kangning and Zeng, Weishuai and Chen, Lipeng and Li, Xi and Wu, Ting and Li, Kailin and Dai, Ruoli and Wang, Jingbo and Han, Lei and Pang, Jiangmiao},
  booktitle={Conference on Robot Learning (CoRL)},
  year={2026}
}

@article{wang2026humanx,
  title={{HumanX}: Toward Agile and Generalizable Humanoid Interaction Skills from Human Videos},
  author={Wang, Yinhuai and Zhao, Qihan and Lau, Yuen Fui and Yu, Runyi and Tsui, Hok Wai and Chen, Qifeng and Wang, Jingbo and Pang, Jiangmiao and Tan, Ping},
  journal={arXiv preprint arXiv:2602.02473},
  year={2026}
}

@inproceedings{ho2020ddpm,
  title={Denoising Diffusion Probabilistic Models},
  author={Ho, Jonathan and Jain, Ajay and Abbeel, Pieter},
  booktitle={Advances in Neural Information Processing Systems (NeurIPS)},
  year={2020}
}

@inproceedings{song2021sde,
  title={Score-Based Generative Modeling through Stochastic Differential Equations},
  author={Song, Yang and Sohl-Dickstein, Jascha and Kingma, Diederik P. and Kumar, Abhishek and Ermon, Stefano and Poole, Ben},
  booktitle={International Conference on Learning Representations (ICLR)},
  year={2021}
}

@article{ho2022cfg,
  title={Classifier-Free Diffusion Guidance},
  author={Ho, Jonathan and Salimans, Tim},
  journal={arXiv preprint arXiv:2207.12598},
  year={2022}
}

@inproceedings{heusel2017fid,
  title={{GANs} Trained by a Two Time-Scale Update Rule Converge to a Local {Nash} Equilibrium},
  author={Heusel, Martin and Ramsauer, Hubert and Unterthiner, Thomas and Nessler, Bernhard and Hochreiter, Sepp},
  booktitle={Advances in Neural Information Processing Systems},
  year={2017}
}

@inproceedings{cheng2024diverse,
  title={Learning Diverse Skills for Local Navigation under Multi-Constraint Optimality},
  author={Cheng, Jin and Vlastelica, Marin and Kolev, Pavel and Li, Chenhao and Martius, Georg},
  booktitle={IEEE International Conference on Robotics and Automation (ICRA)},
  year={2024}
}

@article{radosavovic2024locomotion,
  title={Real-world Humanoid Locomotion with Reinforcement Learning},
  author={Radosavovic, Ilija and Xiao, Tete and Zhang, Bike and Darrell, Trevor and Malik, Jitendra and Sreenath, Koushil},
  journal={Science Robotics},
  volume={9},
  number={89},
  year={2024}
}
\makeatletter\xdef\app@first{\the\c@NAT@ctr}\makeatother
\clearpage
\appendices

\counterwithin{table}{section}
\counterwithin{equation}{section}
\counterwithin{figure}{section}

\renewcommand{\thetable}{\thesection-\arabic{table}}
\renewcommand{\theequation}{\thesection-\arabic{equation}}
\renewcommand{\thefigure}{\thesection-\arabic{figure}}

\section{Motion Data}
\label{app:data}

All motion data is converted to a common format at \SI{50}{\hertz} that contains the root pose, the joint positions, the root and joint velocities, and the poses of all links of the robot.
Source motions are resampled by linear interpolation of positions and joint angles and by spherical interpolation of orientations.
Velocities are recomputed from the resampled motion, and link poses are obtained by forward kinematics.
For the locomotion data, we additionally include a copy of every clip that is mirrored about the sagittal plane.

\subsection{LAFAN1}
We use a publicly available retargeting of LAFAN1~\citep{harvey2020lafan} to the G1~\citep{lafan1retargeted2025}, which is provided at \SI{30}{\hertz}.
From it, we use a single running sequence for the prior comparison.

\subsection{100STYLE}
We retarget the motion capture data of 100STYLE~\citep{mason2022style}, recorded at \SI{60}{\hertz}, to the G1 with GMR~\citep{araujo2025gmr}.
The retargeted motion is smoothed with a Savitzky--Golay filter with a window of \SI{0.1}{\second}.
It leaves the feet floating a few centimeters above the ground during most of a clip.
We therefore shift every frame vertically by the height of the sole of the lower foot, taken as a rolling minimum over \SI{0.25}{\second} and smoothed over time.
Finally, we manually trim the frames at the beginning and at the end of each clip in which the actor walks into and out of the capture volume or stands in a T-pose.
We only use the forward-walking take of each style.
The prior for stylized locomotion is trained on these takes of all 100 styles (\SI{6.7}{\hour}), and a locomotion prior on the take of the \emph{Neutral} style alone (\SI{4}{\minute}); both durations include the mirrored copies.

\subsection{Box Carrying}
We start from the 312 original robot-object clips of OmniRetarget~\citep{yang2025omniretarget}, in which motion capture of humans moving a large box~\citep{li2023omomo} is retargeted to the G1 together with the box.
We keep the clips in which the box is lifted off the ground, manually remove clips in which a foot strikes the box or which contain otherwise undesired motion, and discard clips in which the box tilts by more than \SI{45}{\degree} while it is carried.
This leaves 56 clips from five subjects with a length of \SIrange{2.4}{7.7}{\second} and a total duration of \SI{5}{\minute}.
The pose and the velocity of the box are stored together with the motion of the robot and are used to compute the condition of the prior.

\subsection{Tennis}
The tennis strokes are reconstructed from video.
The source is CalTennis~\citep{demler2026caltennis}, a dataset of more than \SI{50}{\hour} of multi-view video of tennis practice and match play, recorded with static cameras at \SI{60}{\hertz}.
We treat every video as a monocular recording.
We detect persons in frames sampled at \SI{2}{\hertz} and keep the segments of at least \SI{3}{\second} that show a single player in the foreground.
For these segments, we recover the motion of the player with GVHMR~\citep{shen2024gvhmr} and retarget it to the G1 with GMR~\citep{araujo2025gmr}.
Reconstructions with strong foot sliding, jitter, joint-velocity spikes, or ground penetration are discarded automatically, and the remaining clips are reviewed manually.
During the review, the clips are cut into individual strokes, and every stroke is labeled with its contact frame and as a forehand or a backhand.
The resulting dataset contains 500 strokes (285 forehands and 215 backhands) from 7 recording sessions with a total duration of \SI{24}{\minute}.
It is not mirrored, since the racket is held in the right hand.

\subsubsection{Wrist correction}
The racket is rigidly attached to the right hand of the robot with a fixed grip, and the retargeted motion does not ensure that the racket faces the direction of the swing at contact.
We therefore correct the three joints of the right wrist in every stroke and leave all other joints unchanged.
At the contact frame $t^{c}$, we solve for wrist angles $\bm{\theta}^{\ast}$ in two steps.
We first maximize the alignment of the normal of the string bed with the horizontal velocity of the racket.
Among the solutions with this alignment, we then minimize $(\bm{l} \cdot \bm{e}_z)^2 + \mu \norm{\bm{\theta} - \bm{\theta}^{\text{orig}}(t^{c})}^2$, where $\bm{l}$ is the long axis of the racket, $\bm{e}_z$ the vertical direction, and $\mu = 0.2$, which levels the racket as far as this is possible without a large deviation from the original wrist pose.
We blend the correction into the stroke as
\begin{equation}
    \bm{\theta}(t) = \bm{\theta}^{\text{orig}}(t) + \alpha(t) \left( \bm{\theta}^{\ast} - \bm{\theta}^{\text{orig}}(t^{c}) \right) ,
\end{equation}
where $\alpha$ rises from 0 to 1 before the contact and decays back to 0 after it, in both cases along a quintic polynomial with vanishing first and second derivatives at its ends.
The rise spans \SI{60}{\percent} of the frames before the contact, at most 60 frames, and the decay spans \SI{40}{\percent} of the rise, at most 25 frames.
The corrected motion therefore joins the original motion smoothly, and since the derivative of $\alpha$ is zero at the contact, the wrist velocity at the contact is unchanged.
The condition of the prior is computed from the corrected motion.

\section{Motion Priors}
\label{app:priors}

One prior is trained per task on the data listed in \Cref{tab:tasks}.
All priors share the architecture of \Cref{sec:prior} and the hyperparameters in \Cref{tab:prior_hparams}; they differ only in the data and in the dimension of the condition.

\begin{table}[t]
    \caption{Hyperparameters of the motion priors.}
    \label{tab:prior_hparams}
    \centering
    \footnotesize
    \begin{tabular}{ll}
        \thickhline
        Window length $H$ / frame dimension $D$ & 10 / 59 \\
        Token width & 128 \\
        Transformer blocks / attention heads & 2 / 4 \\
        Feed-forward width & 512 \\
        Parameters & $\approx$ 0.7M \\
        \hline
        Loss & $L_1$, \Cref{eq:fm_loss} \\
        Flow time distribution & $\mathcal{U}[0, 1]$ \\
        Optimizer & AdamW \\
        Learning rate / weight decay & $3 \times 10^{-4}$ / $10^{-4}$ \\
        Gradient clipping (norm) & 1.0 \\
        Batch size & 1024 \\
        Iterations & 200k \\
        Weight averaging (EMA) & none \\
        \hline
        Integration steps for generation & 20 \\
        \thickhline
    \end{tabular}
\end{table}

\subsubsection{Motion features}
\Cref{tab:features} lists the features of a frame and the six groups of the style reward.
Each feature dimension is normalized to $[-1,1]$ by an affine map that sends the 1st and 99th percentiles of the motion data to $-1$ and $1$.

\begin{table}[t]
    \caption{Per-frame motion features.}
    \label{tab:features}
    \centering
    \footnotesize
    \setlength{\tabcolsep}{4pt}
    \begin{tabular}{@{}llc@{}}
        \thickhline
        Group $g$ & Content & Dim. \\
        \hline
        Root position & horizontal offset to last frame, height & 3 \\
        Root orientation & heading-relative, 6D & 6 \\
        Joint positions & all actuated joints & 29 \\
        Key-body positions & feet, torso, wrists, relative to root & 15 \\
        Root linear velocity & & 3 \\
        Root angular velocity & & 3 \\
        \thickhline
    \end{tabular}
    \tabnote{Spatial quantities are in the heading frame of the window's last frame.}
\end{table}

\subsubsection{Conditions}
The style label is a one-hot vector over the 100 styles; mirrored clips share the label of their style.
The hitting target and the object target are computed for every training window at its last frame, in the heading frame of the root.
For the hitting target, the contact point is the center of the string bed at the annotated contact frame, the hitting direction is the normal of the string bed at that frame, a unit vector that is horizontal after the wrist correction, and the time to contact is positive before and negative after the contact.
For the object target, the goal is the final position of the box in the clip; during training of the prior, we perturb it with Gaussian noise with a standard deviation of \SI{5}{\centi\meter} so that the exact value does not identify the clip.
Continuous conditions are standardized with the mean and standard deviation over $20\,000$ training windows.

\section{Relation to Score-Matching Motion Priors}
\label{app:smp}

This appendix details the relation between the style reward of SMP~\citep{mu2025smp} and ours that is summarized in \Cref{sec:smp}.

\subsection{Diffusion Models}
A diffusion model~\citep{ho2020ddpm,song2021sde} defines noise levels $i = 1, \dots, K$ with signal levels $1 > \bar{\alpha}_1 > \dots > \bar{\alpha}_K \approx 0$ and corrupts a data sample to
\begin{equation}
    \mw^{(i)} = \sqrt{\bar{\alpha}_i}\, \mw^1 + \sqrt{1 - \bar{\alpha}_i}\, \noise .
    \label{eq:diffusion_forward}
\end{equation}
A network $\hat{\noise}_\theta$ is trained to predict the noise from this sample,
\begin{equation}
    \mathcal{L}(\theta) = \mathbb{E}_{\mw^1,\; \noise,\; i}\, \big\lVert \hat{\noise}_\theta(\mw^{(i)}, i) - \noise \big\rVert^2 .
    \label{eq:diffusion_loss}
\end{equation}
Samples are generated by starting from pure noise at $i = K$ and removing the predicted noise step by step.
The predicted noise has two further interpretations.
First, it is proportional to an estimate of the score of the distribution $p_i$ of corrupted samples, $\nabla \log p_i(\mw^{(i)}) \approx - \hat{\noise}_\theta(\mw^{(i)}, i) / \sqrt{1 - \bar{\alpha}_i}$, which points toward samples that are more likely under the data~\citep{song2021sde}.
Second, solving \Cref{eq:diffusion_forward} for the data sample turns it into a one-step estimate of the clean sample,
\begin{equation}
    \mwref^{(i)} = \frac{\mw^{(i)} - \sqrt{1 - \bar{\alpha}_i}\; \hat{\noise}_\theta(\mw^{(i)}, i)}{\sqrt{\bar{\alpha}_i}} ,
    \label{eq:clean_estimate}
\end{equation}
which, for an optimal model, is the mean of all data samples that are compatible with $\mw^{(i)}$.

The two model classes are closely related.
The diffusion and flow paths both interpolate between a data and a noise sample and differ only in the schedule of the two coefficients.
A velocity prediction also implies a noise prediction, $\hat{\noise} = \mw^{\tau} - \tau\, \vel(\mw^{\tau}, \tau)$, and inserting it into \Cref{eq:clean_estimate} with $\tau$ and $1 - \tau$ in place of $\sqrt{\bar{\alpha}_i}$ and $\sqrt{1 - \bar{\alpha}_i}$ recovers the same clean estimate.
Both can also be made conditional by passing a condition $\cond$ to the network as an additional input, in which case they model $p(\mw \mid \cond)$.
Conditional diffusion models are often sampled with classifier-free guidance~\citep{ho2022cfg}, which extrapolates from an unconditional toward a conditional prediction.

\subsection{Style Reward of SMP}
SMP noises the policy window to each diffusion step $i$ of a fixed set $\mathcal{K}$ of three steps, $\mw_t^{(i)} = \sqrt{\bar{\alpha}_i}\, \mw_t^1 + \sqrt{1 - \bar{\alpha}_i}\, \noise$, and rewards a small noise-prediction error,
\begin{equation}
    r_t^{\text{SMP}} = \exp\!\Big( - \frac{w_s}{|\mathcal{K}|} \sum_{i \in \mathcal{K}} \frac{1}{\mu_i} \big\lVert \hat{\noise}_\theta(\mw_t^{(i)}, i) - \noise \big\rVert^2 \Big) ,
    \label{eq:smp_reward}
\end{equation}
where $\mu_i$ is a running mean of the error at step $i$.
SMP does not denoise the policy window.
By \Cref{eq:clean_estimate}, however, each error equals $\frac{\bar{\alpha}_i}{1 - \bar{\alpha}_i} \lVert \mwref_t^{(i)} - \mw_t^1 \rVert^2$, where $\mwref_t^{(i)}$ is the one-step estimate of the clean window, so \Cref{eq:smp_reward} can equivalently be read as a tracking reward on one-step estimates at three noise levels.
Our projection with a single step, $N = 1$, is its flow-matching counterpart: the residual $\mwref_t - \mw_t^1 = (1 - \tproj) \big( \vel(\mw_t^{\tproj}, \tproj, \cond_t) - (\mw_t^1 - \noise) \big)$ is the residual of the flow-matching loss.
Both rewards thus measure how far the policy window is from the prior's estimate of its noised version.

For an optimal model, a one-step estimate is the mean of all data windows compatible with the noised window, which can be a blend of several motions rather than a natural motion itself.
\method instead integrates the flow with $N$ steps and ends at a single motion; for an exact model and exact integration, natural windows are mapped to references that are again distributed as the data.

\section{Common Training Setup}
\label{app:setup}

\subsubsection{Simulation and control}
The physics runs at \SI{200}{\hertz} and the policy at \SI{50}{\hertz}.
Self-collisions of the robot are disabled.
The action $\bm{a}_t \in \mathbb{R}^{29}$ is mapped to joint position targets $\bm{q}^{\ast} = \bm{q}^{\text{default}} + \bm{s} \odot \bm{a}_t$, where the scale of each joint is $0.25\, \tau^{\max} / k_p$ with its torque limit $\tau^{\max}$ and proportional gain $k_p$.

\subsubsection{Observations}
One actor frame consists of the base angular velocity (3), the projected gravity (3), the joint positions relative to the default pose (29), the joint velocities (29), the previous action (29), and the task command.
The actor receives the last 5 frames.
The critic receives the last 10 frames of the same quantities without noise, together with the base linear velocity and task-specific privileged quantities.
Observations are normalized with running statistics.

\subsubsection{Randomization}
\Cref{tab:randomization} lists the randomization that is shared by all tasks.

\begin{table}[t]
    \caption{Domain randomization and observation noise.}
    \label{tab:randomization}
    \centering
    \footnotesize
    \begin{tabular}{ll}
        \thickhline
        Friction coefficient & $\mathcal{U}[0.3, 1.5]$ \\
        Link masses & $\times\, \mathcal{U}[0.9, 1.1]$ \\
        Link centers of mass & $\pm$\SI{3}{\centi\meter} per axis \\
        PD gains & $\times\, [0.9, 1.1]$, log-uniform \\
        Joint armature & $\times\, [0.9, 1.1]$, log-uniform \\
        Actuation delay & \SIrange{0}{20}{\milli\second} \\
        Pushes & every \SIrange{5}{10}{\second} \\
        \hline
        Base angular velocity noise & $\pm$\SI{0.2}{\radian\per\second} \\
        Projected gravity noise & $\pm 0.05$ \\
        Joint position noise & $\pm$\SI{0.01}{\radian} \\
        Joint velocity noise & $\pm$\SI{1.5}{\radian\per\second} \\
        Joint encoder offset (per episode) & $\pm$\SI{0.015}{\radian} \\
        \thickhline
    \end{tabular}
    \tabnote{Noise is uniform within the given bounds.}
\end{table}

\subsubsection{Termination}
An episode terminates when a body other than the feet and the lower legs touches the ground or, in all tasks except box carrying, when the pelvis drops below \SI{0.3}{\meter}.
In box carrying, the wrists and hands may also touch the ground, and additional conditions apply.
Episodes last \SI{20}{\second}, except in box carrying, where they last \SI{5}{\second}.

\subsubsection{Generative state initialization}
A motion window is generated by integrating the flow from noise with $N=20$ integration steps.
Inverting the canonicalization recovers the root pose, the root velocity, and the joint positions of its last frame; joint velocities are obtained by finite differences, and the robot is reset to this state.
For a conditional prior, we first sample the task command, compute the corresponding condition for the initial configuration, generate the window from $p(\mw \mid \cond)$, and place task objects consistently with it, for example the box at the sampled position relative to the root.
For the unconditional priors, we keep a pool of 4096 generated windows, from which initial states are drawn uniformly, and replace 1024 of them every 2400 control steps.
For the conditional priors, windows are generated on demand in batches of 1024 for the conditions of the upcoming resets.

\subsubsection{Policy optimization}
\Cref{tab:ppo} lists the PPO settings.

\begin{table}[t]
    \caption{PPO hyperparameters.}
    \label{tab:ppo}
    \centering
    \footnotesize
    \begin{tabular}{ll}
        \thickhline
        Environments & 4096 \\
        Steps per env.\ per iteration & 24 \\
        Epochs / mini-batches & 5 / 4 \\
        Discount $\gamma$ / GAE $\lambda$ & 0.99 / 0.95 \\
        Clipping range (policy and value) & 0.2 \\
        Learning rate & $10^{-3}$, adaptive in $[10^{-4}, 10^{-2}]$ \\
        Target KL divergence & 0.01 (box carrying: 0.02) \\
        Entropy coefficient & see \Cref{tab:task_hparams} \\
        Value loss coefficient & 1.0 \\
        Gradient clipping (norm) & 1.0 \\
        Actor and critic networks & MLP, 512-256-128, ELU \\
        Action distribution & Gaussian, learned std.\ (init.\ 1.0) \\
        Iterations & $30\,000$ \\
        \thickhline
    \end{tabular}
\end{table}

\section{Task Details}
\label{app:tasks}

\Cref{tab:task_hparams} lists the parameters of the style reward and of the reward combination for each task.
The group weights are $w_g = 1$ and the scale $\bar{e}$ is frozen after $5 \times 10^{7}$ samples in all tasks, which corresponds to roughly 500 policy iterations.
The action-rate regularizer is $r^{\text{reg}} = \exp(-\norm{\bm{a}_t - \bm{a}_{t-1}}^2 / \sigma_a^2)$.

\begin{table}[t]
    \caption{Task-specific hyperparameters.}
    \label{tab:task_hparams}
    \centering
    \footnotesize
    \setlength{\tabcolsep}{3.5pt}
    \begin{tabular}{@{}lccccc@{}}
        \thickhline
        & \multicolumn{3}{c}{Locomotion} & & \\
        \cmidrule(lr){2-4}
        & Goal & Comp.$^{\dagger}$ & Stylized & Box & Tennis \\
        \hline
        Projection time $\tproj$ & 0.8 & 0.5 & 0.4 & 0.8 & 0.4 \\
        Projection steps $N$ & 4 & 4 & 4 & 4 & 4 \\
        Sharpness $\kappa$ & 6 & 6 & 8 & 2 & 8 \\
        Reward combination$^{\ddagger}$ & G & G & G & M & M \\
        Floor $\beta$ & 0.3 & 0.1 & 0.3 & -- & -- \\
        Weight $\lambda_{\text{task}}$ & -- & -- & -- & 1 & 1 \\
        Weight $\lambda_{\text{reg}}$ & -- & -- & -- & 0.5 & 0.3 \\
        Weight $\lambda_{\text{style}}$ & -- & -- & -- & $0.5 \rightarrow 1$ & 0.7 \\
        Action-rate scale $\sigma_a$ & 5 & 5 & 4 & 2 & 4 \\
        Entropy coefficient & 0.002 & 0.002 & 0.002 & 0.005 & 0.004 \\
        Condition dimension & -- & -- & 100 & 5 & 7 \\
        Command dimension & 2 & 2 / 5 & 2 + 14 & 5 & 12 \\
        \thickhline
    \end{tabular}
    \tabnote{$^{\dagger}$Goal reaching and steering in the prior comparison. $^{\ddagger}$G: gated (\Cref{eq:gated_reward}), M: multi-critic (\Cref{eq:multi_critic}). The weights $\lambda_i$ apply to the multi-critic combination only and the floor $\beta$ to the gated combination only; the command dimension is that of the actor.}
\end{table}

\subsection{Locomotion with a Goal Command}
The goal is placed in a uniformly random direction from the current position of the robot at a distance drawn from $\mathcal{U}[1, 10]\,\si{\meter}$ ($\mathcal{U}[1, 5]\,\si{\meter}$ for stylized locomotion) and is resampled every \SIrange{5}{10}{\second}.
The policy observes the horizontal offset to the goal in its heading frame.
With the horizontal positions of the goal $\bm{p}^{\text{goal}}$ and of the root $\bm{p}$, the task reward is
\begin{equation}
    r^{\text{task}} = \exp\!\left(-0.25\, \norm{\bm{p}^{\text{goal}} - \bm{p}}\right) .
\end{equation}

\subsection{Locomotion with a Steering Command}
As in SMP~\citep{mu2025smp}, the command consists of a travel direction $\bm{d}^{\ast}$ and a facing direction $\bm{f}^{\ast}$, both unit vectors in the horizontal plane, and a speed.
We sample the speed as $v^{\ast} \sim \mathcal{U}[0, 2]\,\si{\meter\per\second}$ and set speeds below \SI{0.5}{\meter\per\second} to zero, so that a quarter of the commands ask the robot to stand still.
Directions are drawn uniformly, and the command is resampled every \SIrange{3}{8}{\second}.
The policy observes both directions and the speed in its heading frame.
With the horizontal base velocity $\bm{v}_{xy}$ and the heading direction $\bm{h}$ of the robot, the task reward is
\begin{equation}
    r^{\text{task}} = 0.6\, e^{-\norm{\bm{v}_{xy} - v^{\ast} \bm{d}^{\ast}}^2}\, \mathbb{1}[\bm{v}_{xy} \cdot \bm{d}^{\ast} \geq 0] + 0.4\, \max(0, \bm{h} \cdot \bm{f}^{\ast}) .
\end{equation}

\subsection{Stylized Locomotion}
The goal and the task reward are those of locomotion with a goal command.
The style command is drawn from 14 of the 100 styles (\emph{Aeroplane}, \emph{ArmsAboveHead}, \emph{BentKnees}, \emph{Chicken}, \emph{Dinosaur}, \emph{DuckFoot}, \emph{HandsBetweenLegs}, \emph{HighKnees}, \emph{LegsApart}, \emph{Neutral}, \emph{ShieldedLeft}, \emph{Star}, \emph{Superman}, \emph{Zombie}) and switches to a different style every \SIrange{5}{10}{\second}.
The actor and the critic observe it as a one-hot vector over these 14 styles, while the prior receives the label over all 100 styles.
Styles are sampled adaptively such that styles with a low style reward are practiced more often: with an exponential moving average $\bar{r}_s$ of the style reward obtained under style $s$ (rate 0.01), the sampling probability is
\begin{equation}
    P(s) = (1 - \eta)\, \frac{\exp(-\bar{r}_s / T)}{\sum_{s'} \exp(-\bar{r}_{s'} / T)} + \frac{\eta}{14} ,
\end{equation}
with $T = 0.1$ and $\eta = 0.1$.
Initial states are generated from the prior conditioned on the initial style.

\subsection{Box Carrying}
\label{app:tasks_box}
\subsubsection{Scene}
The box has a nominal size of $0.32 \times 0.34 \times 0.36\,\si{\meter}$.
Per environment, we scale each of its dimensions by $\mathcal{U}[0.8, 1.2]$, draw its mass from $\mathcal{U}[0.2, 1.0]\,\si{\kilogram}$ and its friction coefficient from $\mathcal{U}[1.0, 1.6]$, and shift its center of mass by up to \SI{2}{\centi\meter} per axis.
The goal lies \SIrange{0.8}{2.0}{\meter} from the initial position of the box, in a direction within $\pm \SI{120}{\degree}$ of the initial heading of the robot.
The actor observes the condition $\cond_t$ with uniform noise of $\pm$\SI{2}{\centi\meter}; the critic additionally observes the stage defined below and the size of the box.

\subsubsection{Initial states}
To expose the policy to all phases of the task, episodes start with the box on the ground in front of the robot (\SI{35}{\percent}), partially lifted (\SI{20}{\percent}), carried on the way to the goal (\SI{30}{\percent}), or held above the goal (\SI{15}{\percent}).
In each case, the initial state of the robot is generated from the prior conditioned on the corresponding relative position of the box, and the box is then placed at this position relative to the root.

\subsubsection{Task reward}
Let $h$ be the height of the box center above its resting height, $d$ its horizontal distance to the goal, and $\bm{\omega}^{\text{box}}$ its angular velocity.
The task proceeds through stages $k \in \{0, 1, 2, 3\}$ that can only advance: the box is lifted ($k = 1$) once $h \geq \SI{0.4}{\meter}$, has reached the goal ($k = 2$) once it is additionally within $d \leq \SI{0.5}{\meter}$, and is placed ($k = 3$) once it rests on the ground within this radius.
The task reward is
\begin{equation}
    r^{\text{task}} = 2\, \frac{k + \rho_k(h)}{4} + 0.4\, e^{-d^2} + 0.1\, e^{-\norm{\bm{\omega}^{\text{box}}}^2 / 9} ,
\end{equation}
with $d$ in meters, $\bm{\omega}^{\text{box}}$ in \si{\radian\per\second}, $\rho_0(h) = \rho_1(h) = \exp(-((h - 0.5) / 0.5)^2)$, which rewards lifting the box to and holding it at a carrying height of \SI{0.5}{\meter}, and $\rho_2(h) = \rho_3(h) = \exp(-(h / 0.5)^2)$, which rewards lowering it.
The regularizer is the product of the action-rate term and $\exp(-(\dot{q}_{\text{rms}} / 10)^2)$, where $\dot{q}_{\text{rms}}$ is the root mean square of the joint velocities in \si{\radian\per\second}.
The weight of the style channel is increased linearly from 0.5 to 1.0 between iterations $10\,000$ and $20\,000$.

\subsubsection{Termination}
In addition to the ground-contact condition, an episode terminates when the box is dropped, that is, when it lands with a speed above \SI{1}{\meter\per\second}, when the robot presses the box against the ground with its torso or hips, or when the state leaves the range covered by the motion data.
For the last condition, we compute per-dimension bounds from the 0.5th and 99.5th percentiles of the data, with a margin, for the pelvis height, the projected gravity, the joint positions, and the position of the box relative to the robot.
Outside of these bounds, the prior is queried with states and conditions that it has not been trained on, and its reference is not meaningful.

\subsection{Tennis}
\label{app:tasks_tennis}
\subsubsection{Scene}
The court is scaled to 0.75 of a regulation court.
The net is \SI{4.8}{\meter} in front of the origin of the environment and \SI{0.914}{\meter} high.
The ball has a radius of \SI{3.3}{\centi\meter} and a mass of \SI{57}{\gram}.
For each ball, we sample a desired hitting point in a box of $[-0.5, 1.5] \times [-2, 2] \times [0.5, 1.5]\,\si{\meter}$ around the origin, a landing target in $[6, 9.6] \times [-3, 3]\,\si{\meter}$ on the opposite side, and a flight time of \SIrange{1}{2}{\second}.
The launch state of the ball is computed such that it bounces once and passes the hitting point at the apex after the bounce; its velocity after the bounce is perturbed by up to \SI{25}{\percent} per axis.
The next ball is launched \SIrange{0.5}{1.0}{\second} after the scheduled contact.
At a reset, the robot is placed relative to the first hitting point for a forehand or a backhand with equal probability, and its initial state is generated from the prior with a time to contact in $[-0.6, 2.8]\,\si{\second}$, so that episodes start in any phase of a stroke.

\subsubsection{Flight model}
\label{app:flight_model}
The ball is modeled as a point mass under gravity and quadratic air drag,
\begin{equation}
    \dot{\bm{v}} = \bm{g} - k_d \norm{\bm{v}}\, \bm{v} ,
\end{equation}
where $\bm{v}$ is the velocity of the ball, $\bm{g}$ the gravitational acceleration, and $k_d = \SI{0.0212}{\per\meter}$ the drag coefficient; spin is not modeled.
The model is integrated with the fourth-order Runge--Kutta method.
The ball bounces when its center reaches a height of one ball radius above the ground.
The integration step in which this happens is split at the crossing, and the velocity is mapped as
\begin{equation}
    v_z \leftarrow -c_n\, v_z, \qquad \bm{v}_{xy} \leftarrow c_t\, \bm{v}_{xy} ,
\end{equation}
with a normal restitution of $c_n = 0.825$ and a tangential retention of $c_t = 0.588$.
The three constants are fitted to recorded trajectories of the real ball.
The same model is used to compute the launch state of the ball, to predict its trajectory for the command, and to predict the landing point of the returned ball for the reward.

\subsubsection{Command}
The hitting target is not taken from the sampled values but estimated online from the state of the ball.
Every two control steps, the flight model is rolled out from the current position and velocity of the ball over a horizon of \SI{2.5}{\second} with a step of \SI{0.04}{\second}.
Among the predicted states before the second bounce that lie inside the box of hitting points and at least \SI{0.15}{\second} ahead, we select the one that is closest to the desired hitting point in the forward direction and in height, and refine it by quadratic interpolation between the neighboring states.
If no predicted state lies inside the box, the state closest to the box is used, provided that it is within \SI{0.8}{\meter}.
The selected state yields the interception point, the velocity $\bm{v}^{\text{in}}$ of the ball at contact, and the time to contact.
The hitting direction is the horizontal unit vector from the interception point to the landing target.
The hitting target is frozen \SI{0.2}{\second} before the contact.
From the interception point, we compute the velocity $\bm{v}^{\text{out}}$ with which the ball has to leave the racket to cross the net with a margin of \SI{0.2}{\meter} and to land at the target.
It is initialized in closed form with the drag-free parabola through the interception point, the crossing point above the net, and the landing target, and is corrected for drag with one shooting iteration through the flight model.
Without an incoming ball, a fixed ready command is used.
The prior receives the hitting target of \Cref{sec:tasks} as its condition.
In its command, the actor receives $\bm{v}^{\text{out}}$ in place of the hitting direction, as well as the position of the ball and the landing target relative to the robot; the critic additionally observes the velocity of the ball.

\subsubsection{Task reward}
Let $\bm{p}^{s}$ and $\bm{v}^{s}$ be the position and velocity of the center of the string bed, $\bm{n}$ its normal, $\bm{p}^{\text{hit}}$ the interception point, and $\bm{v}^{\text{in}}$ the predicted velocity of the ball at contact.
We model the impact with an impulse along the normal of the string bed and a restitution coefficient $c_r = 0.43$ between ball and racket, for which the desired face normal and normal speed of the racket are
\begin{equation}
    \bm{n}^{\ast} = \frac{\bm{v}^{\text{out}} - \bm{v}^{\text{in}}}{\norm{\bm{v}^{\text{out}} - \bm{v}^{\text{in}}}}, \qquad
    v_n^{\ast} = \frac{(\bm{v}^{\text{out}} + c_r\, \bm{v}^{\text{in}}) \cdot \bm{n}^{\ast}}{1 + c_r} .
\end{equation}
The task reward is
\begin{equation}
    r^{\text{task}} = \exp\!\left(-(t^{\text{hit}} / 0.12)^2\right) r^{\text{pos}}\, r^{\text{face}}\, r^{\text{vel}} + 0.2\, r^{\text{land}} ,
\end{equation}
with $r^{\text{pos}} = \exp(-50 \norm{\bm{p}^{s} - \bm{p}^{\text{hit}}}^2)$, $r^{\text{face}} = \exp(-5\, (1 - |\bm{n} \cdot \bm{n}^{\ast}|))$, and $r^{\text{vel}} = \exp(-0.2\, (\bm{v}^{s} \cdot \bm{n}^{\ast} - v_n^{\ast})^2)$.
The first term is gated in time and is only active within about \SI{0.1}{\second} of the contact.
The landing term is set when the racket strikes the ball to $\exp(-\epsilon_{\text{land}}^2) \exp(-\epsilon_{\text{net}}^2)$, where $\epsilon_{\text{land}}$ is the distance in meters between the predicted landing point of the returned ball and the target and $\epsilon_{\text{net}}$ is the deviation of its height at the net from the desired crossing height.
It is zero if the ball does not clear the net and decays with a time constant of \SI{0.3}{\second} afterwards.

\subsubsection{Two-stage training}
Resolving the contact between ball and racket in the simulator requires a physics rate of \SI{2}{\kilo\hertz}.
We therefore train in two stages.
In the first stage, the physics runs at \SI{200}{\hertz}, the ball follows the flight model, and the strike is computed analytically from the impact model above when the string bed sweeps through the ball.
In the second stage, training continues at \SI{2}{\kilo\hertz} with the ball and its contacts simulated by the physics engine.
The scale $\bar{e}$ of the style reward is kept from the first stage.

\section{Experiment Details}
\label{app:experiments}

\subsection{Task Metrics}
\label{app:metrics}
In tennis, strike is the fraction of launched balls that the racket meets.
Clean hit and commanded stroke are the fractions of hitting commands answered with a clean hit and with a clean hit of the commanded stroke, and the contact error is the distance between the center of the string bed and the interception point at the scheduled contact, averaged over the hitting commands.
Net cleared is the fraction of struck balls whose predicted flight clears the net, and the landing error is the distance of their predicted landing point to the commanded target.
In box carrying, an episode counts as placed when the box has been lifted by \SI{0.4}{\meter}, has been within \SI{0.5}{\meter} of the goal, and has been set down there; the final distance combines the horizontal distance of the box to the goal with its height above the resting height at the end of the episode.
In goal reaching, a goal is reached if the root comes within \SI{0.5}{\meter} of it before it is resampled.

\subsection{Fréchet Distance}
\label{app:frechet}
The Fréchet distance is computed between Gaussians fitted to two sets of motion features, normalized as for the prior.
It is divided by the distance between two disjoint sets of the motion data.
In tennis, box carrying, and the projection experiments, the features are the root rotation, joint positions, and key-body positions of a ten-frame window (500 dimensions), and the windows of the policy or of the projection are compared with windows of the motion data; in tennis, both sets are restricted to windows that end \SIrange{0.4}{0.6}{\second} before contact.
In goal reaching, steering, and stylized locomotion, the features are those of the last frame of the window (50 dimensions) and are compared with held-out validation data, in stylized locomotion per commanded style and averaged over the styles.

\subsection{Baselines}
\label{app:baselines}
All variants of an experiment share the motion data, the task reward, and the observations, and differ in the source of the style reward.
The AMP discriminator~\citep{peng2021amp} sees the same ten-frame windows as our prior, has hidden layers of 512, 256, and 128 units, and is trained with a least-squares loss and a gradient penalty; its reward is $r^{\text{style}} = \max(0, 1 - \frac{1}{4}(D - 1)^2)$.
The SMP prior is a denoising diffusion model with the network and data of the \method prior, 50 diffusion steps with a cosine schedule, and an $L_1$ noise-prediction loss; its reward follows \Cref{eq:smp_reward} with $\mathcal{K} = \{8, 15, 22\}$.
In the conditional variants, the standardized condition is appended to the input of the discriminator and given to the diffusion model; the unconditional priors are trained on the same data without the condition.
Task-only training sets the weight of the style channel to zero or, with the gated reward, $r^{\text{style}} = 1$.
The frozen discriminator of \Cref{tab:prior_comparison} is the final AMP checkpoint of the other task.

\subsection{Evaluation}
\label{app:evaluation}
All policies are evaluated in 1024 environments (1000 in tennis) without observation noise, randomization, or pushes, with identical command schedules for the policies of an experiment.
In tennis and box carrying, every variant starts its episodes from states generated by the conditional prior (\Cref{sec:gsi}), and every box-carrying episode of the evaluation starts with the box on the ground.
The tennis policies are evaluated after the first of the two training stages, in which the impact of the ball on the racket is computed analytically; the conditional \method policies of \Cref{tab:cond} are the $\tproj = 0.4$, $N = 4$ policies of \Cref{sec:exp_design}, of which two seeds play both strokes and the third answers almost all commands with a forehand.

\end{document}